\documentclass{ecai}  

\usepackage{graphicx}
\usepackage{latexsym}

\usepackage{amssymb}
\usepackage{amsthm}
\usepackage{booktabs}
\usepackage[all]{foreign}
\usepackage[hidelinks]{hyperref}
\usepackage{multirow}
\usepackage{subfig}
\usepackage{paralist}
\usepackage[inline]{enumitem}
\usepackage{tikz}

\usepackage{xr}
\usepackage{refcount}

\usetikzlibrary{automata, positioning, arrows, arrows.meta, shapes.geometric}

\tikzstyle{world}=[rectangle,thick,draw=black,fill=black,minimum size=4pt,inner sep=0pt]

\tikzstyle{every initial by arrow}=[initial text=]

\newtheorem{example}{Example}
\newtheorem{thm}{Theorem}
\newtheorem{definition}{Definition} 
\newtheorem{lemma}{Lemma}
\newtheorem{proposition}{Proposition}
\newtheorem{corollary}{Corollary}

\usepackage[table]{colortbl}
\usepackage{xifthen}
\usepackage{mathtools}
\usepackage{tikz-cd}
\usetikzlibrary{decorations.pathmorphing}

\newcommand{\B}[1]{\Box_{#1}}
\newcommand{\D}[1]{\Diamond_{#1}}
\newcommand{\CK}[1]{C_{#1}}

\newcommand{\atom}[1]{#1}

\newcommand{\atomSet}{\mathcal{P}}

\newcommand{\agentSet}{\mathcal{AG}}
\newcommand{\actionSet}{\mathcal{A}}

\newcommand{\E}{\mathcal{E}}
\newcommand{\Lang}[1]{\mathcal{L}_{\atomSet, \agentSet}^{#1}}

\newcommand{\axiom}[1]{\textbf{#1}}
\newcommand{\planex}[2]{\textnormal{\textsc{PlanEx}}(#1, #2)}

\newcommand\boldparagraph[1]{\textbf{\textit{#1}}~}

\newcommand{\METAWORLD}{\textsf{META-WORLD}}
\newcommand{\METACHAIN}{\textsf{META-CHAIN}}
\newcommand{\METAINC}{\textsf{META-INC}}
\newcommand{\METADEC}{\textsf{META-DEC}}
\newcommand{\METAREPL}{\textsf{META-REPL}}
\newcommand{\METASTATE}{\textsf{META-S}}

\tikzstyle{world}       =[circle,   thick,draw=black,       fill=black,minimum size=5pt,inner sep=0pt]
\tikzstyle{pointedworld}=[circle,   thick,draw=black,double,fill=black,minimum size=5pt,inner sep=0pt]
\tikzstyle{event}       =[rectangle,thick,draw=black,       fill=black,minimum size=5pt,inner sep=0pt]
\tikzstyle{pointedevent}=[rectangle,thick,draw=black,double,fill=black,minimum size=5pt,inner sep=0pt]
\tikzstyle{itria}       =[draw,isosceles triangle,shape border rotate=90,inner sep=2pt,outer sep=-0.5pt,anchor=north]

\newcommand\bisim{\underline{\leftrightarrow}}

\begin{document}
    \begin{frontmatter}
        \title{A Semantic Approach to\\ Decidability in Epistemic Planning (Extended Version)}

        \author[A]{\fnms{Alessandro}~\snm{Burigana}\thanks{Corresponding Author. Email: burigana@inf.unibz.it.}}
        \author[B]{\fnms{Paolo}~\snm{Felli}}
        \author[A]{\fnms{Marco}~\snm{Montali}}
        \author[A]{\fnms{Nicolas}~\snm{Troquard}}

        \address[A]{Free University of Bozen-Bolzano}
        \address[B]{University of Bologna}

        \begin{abstract}
    The use of Dynamic Epistemic Logic (DEL) in multi-agent planning has led to a widely adopted action formalism that can handle nondeterminism, partial observability and arbitrary knowledge nesting. As such expressive power comes at the cost of undecidability, several decidable fragments have been isolated, mainly based on syntactic restrictions of the action formalism.
	In this paper, we pursue a novel \emph{semantic approach} to achieve decidability. Namely, rather than imposing syntactical constraints, the semantic approach focuses on the axioms of the logic for epistemic planning. Specifically, we augment the logic of knowledge S5$_n$ and with an interaction axiom called \emph{(knowledge) commutativity}, which controls the ability of agents to unboundedly reason on the knowledge of other agents. We then provide a threefold contribution. First, we show that the resulting epistemic planning problem is decidable. In doing so, we prove that our framework admits a finitary non-fixpoint characterization of common knowledge, which is of independent interest. Second, we study different generalizations of the commutativity axiom, with the goal of obtaining decidability for more expressive fragments of DEL.
	Finally, we show that two well-known epistemic planning systems based on action templates, when interpreted under the setting of knowledge, conform to the commutativity axiom, hence proving their decidability.
\end{abstract}


    \end{frontmatter}

    \section{Introduction} 
    Multi-agent systems find applications in a wide range of settings where the agents need to be able to reason about both the physical world and the \emph{knowledge} that other agents possess---that is, their \emph{epistemic state}.  
    \emph{Epistemic planning} \cite{journals/jancl/Bolander2011} employs the theoretical framework of Dynamic Epistemic Logic (DEL) \cite{book/springer/vanDitmarsch2007} in the context of automated planning. The resulting formalism is able to represent nondeterminism, partial observability and arbitrary knowledge nesting. That is, agents have the power to reason about higher-order knowledge of other agents with no limitations.
    
    Due to the high expressive power of the DEL framework, the \emph{plan existence problem} (see Definition \ref{def:plan_ex_problem}), that asks whether there exists a plan to achieve a goal of interest, is undecidable in general \cite{journals/jancl/Bolander2011}. 
    As a consequence, in the past decade, DEL has been widely studied to obtain (un)decidability and complexity results for fragments of the planning problem.
    A common approach (see~Section~\ref{sec:rel_works}) consists in syntactically restricting the action theory, for instance by limiting the modal depth of the preconditions and postconditions of actions to a certain bound $d$ \cite{conf/ijcai/Bolander2015,conf/ijcai/Charrier2016,journals/ai/Bolander2020}. Nonetheless, the problem remains undecidable even with $d{=}2$ when only \emph{purely epistemic actions} are allowed, and with $d{=}1$ when factual change is involved. 
    This suggests that such syntactic restrictions are too strong in many practical cases, where reasoning about the knowledge of others is required.
    
    For this reason, in this paper we pursue a different strategy that we call \emph{semantic approach}. Namely, rather than imposing syntactical constraints, the semantic approach focuses on the axioms of the logic for epistemic planning. Specifically, we consider the multi-agent logic for knowledge S5$_n$ (where $n$ denotes the number of agents) and we augment it with an interaction axiom, called the \emph{(knowledge) commutativity} axiom (where, as customary, $\B{i} \varphi$ indicates that agent $i$ \emph{knows} that $\varphi$ holds):

    \begin{equation*}
        \begin{array}{lll}
            \axiom{C} & \B{i} \B{j} \varphi \rightarrow \B{j} \B{i} \varphi & \textnormal{(Commutativity)}
        \end{array}
    \end{equation*}
    
    \noindent This axiom imposes a principle of commutativity in the higher-order knowledge across agents. In the resulting logic, which we call C-S5$_n$, while agents have their own distinct individual knowledge, higher-order levels of perspectives of agents \emph{commute}. This assumption is well suited in cooperative planning domains \cite{journals/csur/Torreno2017}, where it is required that agents act and communicate in an observable way, thus making knowledge of agents accessible to others.
    
    We provide a threefold contribution. First, we show that the epistemic plan existence problem in the resulting framework becomes decidable. We do so by proving that the commutativity axiom ensures that the states in the logic C-S5$_n$ are bounded in size, which entails that the search space of the plan existence problem is finite. In doing so, we show that the logic C-S5$_n$ admits a \emph{finitary non-fixpoint} characterization of common knowledge, which is often regarded as a possible solution to paradoxes involving common knowledge (see \cite{journals/synthese/Paternotte11} for an overview).
    
    Second, we investigate the plan existence problem with different generalized principles of commutativity.
    Indeed, although the commutativity axiom is better fitting for tight-knit groups of agents, it may be less suited for representing more loosely organized groups.
    We define suitable generalizations parametrized by fixed integer constants $b{>}1$ and $1 {<} \ell {\leq} n$. The resulting axioms are the following (where $\pi$ is a permutation of the sequence $\langle i_1, \dots i_\ell\rangle$ of agents, as explained in more detail in Section \ref{sec:general-comm}):
    
    \begin{equation*}
        \begin{array}{lll}
            \axiom{C$^b$} & (\B{i} \B{j})^b \varphi \rightarrow (\B{j} \B{i})^b \varphi & \textnormal{($b$-Comm.)} \\
            \axiom{wC$_\ell$} & \B{i_1} \dots \B{i_\ell}\varphi \rightarrow \B{\pi_{i_1}} \dots \B{\pi_{i_\ell}}\varphi & \textnormal{(Weak comm.)}
        \end{array}
    \end{equation*}

    \noindent Concerning axiom \axiom{wC$_\ell$}, we show that the plan existence problem remains decidable for any $1 {<} \ell {\leq} n$. Relating axiom \axiom{C$^b$}, we show that the plan existence problem remains decidable in the presence of two agents ($n{=}2$), for any $b{>}1$. We also show that for any $n{>}2$ and any $b{>}1$ the problem becomes undecidable.

    Finally, we show that the knowledge (\ie S5$_n$) fragment of the well known planning system $m\mathcal{A}^*$ \cite{journals/corr/Baral2015} and the system by Kominis and Geffner \cite{conf/aips/Kominis2015} are captured by our formalism. Thus, we prove the decidability of a fragment $m\mathcal{A}^*$, which was still an open problem, and of the action formalism in \cite{conf/aips/Kominis2015}, confirming their previous results.

    Since the axioms of the logic for epistemic planning lie at the core of the semantic approach, we consider such axioms to define \emph{meaningful states}. In other words, when a certain principle is introduced to be an \emph{axiom} of the logic, an epistemic state is considered to be \emph{meaningful} if and only if such principle is satisfied. Thus, when planning under a logic $L$, we consider a plan to be meaningful and, in turn, valid only if all the states that it visits satisfy the axioms of $L$.
    At the same time, the semantics of the product update of DEL (see Definition \ref{def:update_em}) does not guarantee that the application of an action of a generic logic $L$ to an epistemic state of the same logic necessarily results in an epistemic state that satisfies all axioms of $L$. A well-known example of this phenomenon is found in the widely studied doxastic logic KD45$_n$ \cite{books/mit/Fagin2004}, where the \emph{consistency axiom} \axiom{D} is not guaranteed to be preserved by the product update. Addressing the problem of preservation of axioms after action updates is not trivial. Indeed, in the literature, considerable effort has been spent in developing different techniques to handle the preservation of axiom \axiom{D} \cite{workshop/nrac/Herzig2005,conf/aaai/Son2015}. Analogously to the case of \axiom{D} in KD45$_n$, in our framework axioms \axiom{C}, \axiom{C$^b$} and \axiom{wC$_\ell$} are not guaranteed to be preserved by the product update. As a result, as explained above, we consider a plan to be valid if it only visits meaningful epistemic states, \ie those satisfying the axioms of the considered logic. Importantly, our decidability results continue to hold even when one adopts more sophisticated revision techniques that handle the preservation problem by accepting and suitably curating non-preserving states. The development of such non-trivial techniques for our logics is independent of the analysis of decidability of the plan existence problem under the same logics, and is left as an important, follow-up work.
    
    The paper is organised as follows.
    In Section~\ref{sec:del}, we recall some preliminaries and define epistemic planning tasks. In Section~\ref{sec:new_logic}, we discuss in more detail the semantic approach, we introduce our new logic for epistemic planning and we discuss the commutativity axiom. In Section~\ref{sec:decidability}, we analyze decidability of epistemic planning under commutativity and its generalizations. In Section~\ref{sec:systems}, we apply our decidability results to existing epistemic planning systems.
    Finally, in Section~\ref{sec:rel_works}, we discuss related work.

    \section{Dynamic Epistemic Logic}\label{sec:del}
    This section is organized as follows. The syntax and semantics of DEL~\cite{book/springer/vanDitmarsch2007} are introduced in Section~\ref{sec:epistemic_models}, event models and the product update in Section~\ref{sec:event_models}.
    In Section \ref{sec:s5}, we recall the axioms of the logic S5$_n$.
    In Section~\ref{sec:plan_ex} we define the plan existence problem.

\subsection{Epistemic Models}\label{sec:epistemic_models}
    Let $\atomSet$ be a finite set of atomic propositions and $\agentSet = \{1, \dots, n\} $ a finite set of agents. The language $\Lang{C}$ of \emph{multi-agent epistemic logic on $ \atomSet $ and $ \agentSet $ with common knowledge} is defined by the following BNF:
    \begin{equation*}
        \varphi ::= \atom{p} \mid \neg \varphi \mid \varphi \wedge \varphi \mid \B{i} \varphi \mid \CK{G} \varphi,
    \end{equation*}

    \noindent where $ \atom{p} \in \atomSet $, $ i \in \agentSet $, and $ \varnothing \neq G \subseteq \agentSet $. Formulae $ \B{i} \varphi $ and $ \CK{G} \varphi $ are respectively read as ``agent $ i $ knows that $ \varphi $'' and ``group $G$ has common knowledge that $ \varphi $''.
    We define $\top$, $\bot$, $\vee$, $\rightarrow$ and $\D{i}$ as usual.

    \begin{definition}[Epistemic Model and State]
        An \emph{epistemic model} of $ \Lang{C} $ is a triple $ M = (W, R, V) $ where:
        \begin{compactitem}
            \item $ W \neq \varnothing $ is a finite set of possible worlds; 
            \item $ R: \agentSet \rightarrow 2^{W \times W} $ assigns to each agent $ i $ an accessibility relation $R(i)$ (abbreviated as $R_i$);
            \item $ V: \atomSet \rightarrow 2^W $ assigns to each atom a set of worlds.
        \end{compactitem}
        \noindent An \emph{epistemic state} is a pair $ (M, W_d) $, where $ W_d \subseteq W $ is a non-empty set of designated worlds.
    \end{definition}

    \noindent Intuitively, a designated world in $W_d$ is considered the current ``real" world from the perspective of an external observer (the planner) rather than of agents in $\agentSet$. Thus, $|W_d|>1$ represents the uncertainty of the observer about the real world. 
    
    The pair $(W, R)$ is called the \emph{frame} of $M$. We use the infix notation $ w R_i v $ in place of $ (w, v) \in R_i $.
    We also define $ R_G \doteq \cup_{i \in G} R_i $, where $ G \subseteq \agentSet $. The reflexive and transitive closure of $ R $ is denoted by $ R^* $. 
    Relations $R_i$ capture what agents consider to be possible: $ w R_i v $ denotes the fact that, in $w$, agent $i$ considers $v$ to be possible.
    Throughout the paper, to support our exposition, we consider the example of the \emph{coordinated attack problem} \cite{book/springer/Gray1978,journals/jacm/Halpern1990}. It is a well-known problem that is often analyzed in the distributed systems literature. In what follows, we appeal to the DEL representation of this problem provided in \cite{journals/ai/Bolander2020}.

    \begin{example}[The Coordinated Attack Problem]\label{ex:k_model}
        Two generals, $\mathbf{a}$ and $\mathbf{b}$, are camped with their armies on two hilltops overlooking a common valley, where the enemy is stationed. The only way for them to defeat the enemy is to attack simultaneously. They can only communicate by means of a messenger, who may be captured at any time when crossing the valley. Neither general will attack until he is sure that the other will attack as well.
        
        General $\mathbf{a}$ and the messenger are initially together, and general $\mathbf{a}$ decides to attack at dawn. We use the atomic propositions $d$ to denote that `\emph{general $\mathbf{a}$ will attack at dawn}' and $m_i$, for $i = \mathbf{a}, \mathbf{b}$, to denote that the messenger is currently at the camp of general $i$. In this way, $\neg m_a \land \neg m_b$ expresses the fact that the messenger has been captured.
        
        The initial situation can be described by the epistemic state $s_0$ shown below\footnote{In all figures, the reflexive, symmetric and transitive closures of the relations are left implicit.}. Each bullet represents a world and the designated world is denoted by a circled bullet. There are two possible worlds, denoting the possibility that general $\mathbf{a}$ will attack at dawn ($w_1$), or not ($w_2$). Both generals know that the messenger is camped with general $\mathbf{a}$. The fact that general $\mathbf{b}$ does not know whether general $\mathbf{a}$ has decided to attack is represented by the indistinguishability relations between worlds $w_1$ and $w_2$. In fact, initially, general $\mathbf{b}$ has not enough information to know whether his ally has decided to attack.

        {\centering
            \begin{tikzpicture}[-,>=stealth',shorten >=1pt,auto,semithick]
    \node (A0) []                 {$s_0$} ;
    \node (A1) [right=.1cm of A0] {$=$} ;
    \node (W1) [pointedworld, right=.3cm of A1, label=below:{$w_1 : d, m_a$}] {} ;
    \node (W2) [world,        right=2cm of W1,  label=below:{$w_2 : m_a$}] {} ;

    \path (W1)
        edge node [above] {$b$} (W2);
\end{tikzpicture}

        \par}
    \end{example}

    \begin{definition}[Truth in epistemic states]
        Let $ M = (W, R, V) $ be an epistemic model, $ w \in W $, $i \in \agentSet$, $\varnothing \neq G \subseteq \agentSet$, $p \in \atomSet$ and $\varphi,\psi \in \Lang{C}$ be two formulae. Then,
        
        {\centering
        $
            \begin{array}{@{}lll}                    
                (M, w) \models \atom{p}            & \text{ iff } & w \in V(\atom{p}) \\
                (M, w) \models \neg \varphi        & \text{ iff } & (M, w) \not\models \varphi \\
                (M, w) \models \varphi \wedge \psi & \text{ iff } & (M, w) \models \varphi \text{ and } (M, w) \models \psi \\
                (M, w) \models \B{i} \varphi       & \text{ iff } & \forall v \text{ if } w R_i v \text{ then } (M, v) \models \varphi \\
                (M, w) \models \CK{G} \varphi       & \text{ iff } & \forall v \text{ if } w R_G^* v \text{ then } (M, v) \models \varphi
            \end{array}
        $
        \par}

        \noindent Let $(M, W_d)$ be an epistemic state. Then,

        {\centering
        $
            \begin{array}{@{}lll}                    
                (M, W_d) \models \varphi           & \text{ iff } & (M, w) \models \varphi \textnormal{ for all } w \in W_d
            \end{array}
        $
        \par}
    \end{definition}
	
	\noindent For instance, $(M,w)\models p$ means that $p$ is true in $w$; $(M,w)\models \D{i} p$ means that in $w$ the agent $i$ admits the possibility of $\varphi$ being true, i.e., there exists a world $v$ that $i$ considers possible (i.e., with $w R_i v$) such that $(M,v)\models \varphi$; $(M,w)\models \B{i} \varphi$ means that in $w$ the agent $i$ knows $\varphi$, as $\varphi$ holds in all worlds that $i$ considers possible.
	
    We recall the notion of bisimulation for epistemic states \cite{conf/kr/Bolander2021}. 
                
    \begin{definition}[Bisimulation]
        Let $s=((W, R, V), W_d)$ and $s'=((W', R', V'), W'_d)$ be two epistemic states. We say that $s$ and $s'$ are \emph{bisimilar}, denoted by $s \bisim s'$, if there exists non-empty binary relation $ Z \subseteq W \times W' $ satisfying:

        \begin{compactitem}
            \item \emph{Atoms}: if $(w, w') \in Z$, then for all $ \atom{p} \in \atomSet $, $ w \in V(\atom{p}) $ iff $ w' \in V'(\atom{p}) $.
            \item \emph{Forth}: if $(w, w') \in Z$ and $ w R_i v $, then there exists $ v' \in W' $ such that $ w' R'_i v' $ and $ (v, v') \in Z $.
            \item \emph{Back}: if $(w, w') \in Z$ and $ w' R'_i v' $, then there exists $ v \in W $ such that $ w R_i v $ and $ (v, v') \in Z $.
            \item \emph{Designated}: if $ w \in W_d $, then there exists $w' \in W'_d$ such that $(w, w') \in Z$, and vice versa.
        \end{compactitem}

        \noindent We say that $Z$ is a \emph{bisimulation} between $s$ and $s'$.
    \end{definition}

    \noindent Throughout the rest of the paper, we assume that \emph{each epistemic state is minimal w.r.t. bisimulation}.
    We denote the fact that $(w, w') \in Z$ by $w \bisim w'$.
    Finally, we introduce a notion of $k$-bisimulation for epistemic states. The following definition follows the one in \cite{books/cup/Blackburn2001} and generalizes that of \cite{conf/ijcai/Yu2013} by considering epistemic states with (possibly) multiple designated worlds.

    \begin{definition}[$k$-bisimulation]\label{def:k-bisim}
        Let $k \geq 0$ and let $s=((W, R, V),$ $W_d)$ and $s'=((W', R', V'), W'_d)$ be two epistemic states. We say that $s$ and $s'$ are $k$-bisimilar, denoted by $s \bisim_k s'$, if there exists a sequence of non-empty binary relations $Z_k \subseteq \ldots \subseteq Z_0$ (with $Z_0 \subseteq W \times W'$) satisfying (for any $i < k$):
        \begin{compactitem}
            \item \emph{Atoms}: if $(v, v') \in Z_0$, then for all $ \atom{p} \in \atomSet $, $ v {\in} V(\atom{p}) $ iff $ v' {\in} V'(\atom{p}) $.
            \item \emph{Forth}: if $(v, v') \in Z_{i+1}$ and $ v R_i u $, then there exists $ u' \in W' $ such that $ v' R'_i u' $ and $ (u, u') \in Z_i $.
            \item \emph{Back}: if $(v, v') \in Z_{i+1}$ and $ v' R'_i u' $, then there exists $ u \in W $ such that $ v R_i u $ and $ (u, u') \in Z_i $.
            \item \emph{Designated}: if $ v \in W_d $, then there exists $v' \in W'_d$ such that $(v, v') \in Z_k$, and vice versa.
        \end{compactitem}

        \noindent We say that $Z_k$ is a \emph{$k$-bisimulation} between $s$ and $s'$.
    \end{definition}

    \subsection{Event Models and Product Update}\label{sec:event_models}
        Information change is captured by \emph{product updates} of the current epistemic state with the \emph{event model} of actions. 
        
        \begin{definition}[Event Model and Action]
            An \emph{event model} for $ \Lang{C} $ is a tuple $ \E = (E, Q, pre, post) $ where:
            \begin{compactitem}
                \item $ E \neq \varnothing $ is a finite set of events; 
                \item $ Q: \agentSet \rightarrow 2^{E \times E} $ assigns to each agent $ i $ an accessibility relation $ Q(i) $ (abbreviated as $Q_i$); 
                \item $ pre: E \rightarrow \Lang{C} $ assigns to each event a \emph{precondition}; 
                \item $ post: E \rightarrow (\atomSet \rightarrow \Lang{C}) $ assigns to each event and atom a \emph{postcondition}.
            \end{compactitem}
            \noindent An \emph{action} is a pair $ (\E, E_d) $ where $ E_d \subseteq E $ is a non-empty set of designated events.
        \end{definition}

        \noindent Similarly to epistemic states, the designated events in $E_d$ represent the ``real'' events that are taking place from the perspective of an external observer.

        The pair $(E, Q)$ is called the \emph{frame} of $\E$.
 		We use the infix notation $ e Q_i f $ in place of $ (e, f) \in Q_i$. 
 		These relations are analogous to the accessibility relations of epistemic models: they are used to specify how the knowledge of each agent is affected by an action, depending on which events each agent considers possible. 
		Intuitively, the precondition of an event $e$ specify whether $e$ \emph{could happen} in a certain world $w$, whereas the postconditions of $e$ describe how such event might change the factual properties of $w$ (see Definition \ref{def:update_em}). Formally, we say that an event $e$ is \emph{applicable} in a world $w$ of $M$ if $(M,w)\models pre(e)$.
        
        \begin{example}\label{ex:e_model}
            Imagine that general $\mathbf{a}$ decides to send the messenger to general $\mathbf{b}$ (action send$_{ab}$). While doing  so, the general considers two possible outcomes:
            \begin{enumerate*}[label=\arabic*)]
                \item the messenger safely arrives to the other side of the valley, or
                \item the messenger is captured by the enemy. 
            \end{enumerate*}
            In the figure below, these eventualities are represented by events $e^a_1$ and $e^a_2$, respectively. The precondition of $e^a_1$ is $pre(e^a_1) = d \land m_a$, namely the message can only arrive to general $\mathbf{b}$ if general $\mathbf{a}$ has indeed decided to attack at dawn and if the messenger is at $\mathbf{a}$'s camp. The precondition of $e^a_2$ is simply $pre(e^a_2) = \top$, since the messenger could always be captured. We represent the fact that the messenger travels from one hilltop to the other\footnote{For simplicity, we assume that the truth value of each atomic proposition remains unchanged unless explicitly specified.} by having $post(e^a_1)(m_a) = \bot$ and $post(e^a_1)(m_b) = \top$. Finally, we denote the fact that the messenger is captured by having $post(e^a_2)(m_a) = \bot$ and $post(e^a_2)(m_b) = \bot$.

            {\centering
                \begin{tikzpicture}[-,>=stealth',shorten >=1pt,auto,semithick]
    \node (A0) []                 {send$_{ab}$} ;
    \node (A1) [right=.1cm of A0] {$=$} ;
    \node (E1) [pointedevent, right=.3cm of A1, label=below:{$e^a_1$}] {} ;
    \node (E2) [event,        right=2cm of E1, label=below:{$e^a_2$}] {} ;

    \path (E1)
        edge node [above] {$a$} (E2);
\end{tikzpicture}

            \par}

            Action send$_{ba}$ is defined similarly, by having $pre(e^b_1) = d \land m_b$, $pre(e^b_2) = \top$, $post(e^b_1)(m_a) = \top$, $post(e^b_1)(m_b) = \bot$, $post(e^b_2)(m_a) = \bot$ and $post(e^b_2)(m_b) = \bot$.

            {\centering
                \begin{tikzpicture}[-,>=stealth',shorten >=1pt,auto,semithick]
    \node (A0) []                 {send$_{ba}$} ;
    \node (A1) [right=.1cm of A0] {$=$} ;
    \node (E1) [pointedevent, right=.3cm of A1, label=below:{$e^b_1$}] {} ;
    \node (E2) [event,        right=2cm of E1, label=below:{$e^b_2$}] {} ;

    \path (E1)
        edge node [above] {$b$} (E2);
\end{tikzpicture}

            \par}
        \end{example}
        
        The \emph{product update} formalizes the execution of an action $(\E, E_d)$ on the current epistemic state $(M, W_d)$. Intuitively, the resulting epistemic state $(M', W_d')$ is computed by a cross product between the worlds in $M$ and the events in $\E$. A pair $(w,e)$ represents the world of $M'$ that results from applying the event $e$ on the world $w$.
        We say that $ (\E, E_d) $ is \emph{applicable} in $ (M, W_d) $ iff for each world $ w_d \in W_d $ there exists an event $e_d\in E_d$ that is applicable in $w_d$. 
        
        \begin{definition}[Product Update]\label{def:update_em}
            Let $ (\E, E_d) $ be an action applicable in an epistemic state $ (M, W_d) $, where $ M = (W, R, V) $ and $ \E = (E, Q, pre, post) $. The \emph{product update} of $ (M, W_d) $ with $ (\E, E_d) $ is the epistemic state $ (M, W_d) \otimes (\E, E_d) = ((W', R', V'), W'_d) $, where:

            {\centering
            $
                \begin{array}{l@{\;}ll}
                    W'	         & = \{(w, e) \in W {\times} E \mid (M, w) \models pre(e)\} \\
                    R'_i	     & = \{((w, e), (v, f)) \in W' {\times} W' \mid w R_i v \text{ and } e Q_i f\} \\
                    V'(\atom{p}) & = \{(w, e) \in W' \mid (M, w) \models post(e)(\atom{p})\} \\
                    W'_d	     & = \{(w, e) \in W' \mid w \in W_d \text{ and } e \in E_d\}
                \end{array}
            $
            \par}
        \end{definition}

        \begin{example}\label{ex:update}
            Suppose that general $\mathbf{a}$ sends the messenger to general $\mathbf{b}$ (action send$_{ab}$) and that the message is successfully delivered. The situation is represented by epistemic state $s_1$, where $w'_1 = (w_1, e^a_1)$, $w'_2 = (w_1, e^a_2)$ and $w'_3 = (w_2, e^a_2)$ (recall that the reflexive, symmetric and transitive closures of the relations are left implicit).

            {\centering
                \begin{tikzpicture}[-,>=stealth',shorten >=1pt,auto,semithick]
    \node (A0) []                 {$s_1$} ;
    \node (A1) [right=.1cm of A0] {$=$} ;
    \node (W1) [pointedworld, right=.3cm of A1,  label=below:{$w'_1 : d, m_b$}] {} ;
    \node (W2) [world,        right=1.5cm of W1, label=below:{$w'_2 : d$}] {} ;
    \node (W3) [world,        right=1.5cm of W2, label=below:{$w'_3 $}] {} ;

    \path (W1)
        edge node [above] {$a$} (W2) ;
    
    \path (W2)
        edge node [above] {$b$} (W3) ;
\end{tikzpicture}

            \par}
            
            Now general $\mathbf{b}$ knows about the intentions of his ally ($\B{b}d$), but general $\mathbf{a}$ does not know that general $\mathbf{b}$ knows. So, $\mathbf{b}$ decides to send the messenger back to acknowledge that the message was received (action send$_{ba}$). Here, $d$ assumes the meaning of ``general $\mathbf{b}$ has received the message''. Assume again that the messenger succeeds. We obtain the epistemic state $s_2$, where $w''_1 = (w'_1, e^b_1)$, $w''_2 = (w'_1, e^b_2)$, $w''_3 = (w'_2, e^b_2)$ and $w''_4 = (w'_3, e^b_2)$.
            
            
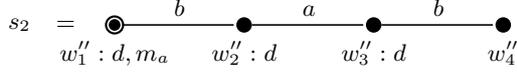
{\centering
                \begin{tikzpicture}[-,>=stealth',shorten >=1pt,auto,semithick]
    \node (A0) []                 {$s_2$} ;
    \node (A1) [right=.1cm of A0] {$=$} ;
    \node (W1) [pointedworld, right=.3cm of A1,  label=below:{$w''_1 : d, m_a$}] {} ;
    \node (W2) [world,        right=1.5cm of W1, label=below:{$w''_2 : d$}] {} ;
    \node (W3) [world,        right=1.5cm of W2, label=below:{$w''_3 : d$}] {} ;
    \node (W4) [world,        right=1.5cm of W3, label=below:{$w''_4 $}] {} ;

    \path (W1)
        edge node [above] {$b$} (W2) ;
    
    \path (W2)
        edge node [above] {$a$} (W3) ;

    \path (W3)
        edge node [above] {$b$} (W4) ;
    
\end{tikzpicture}

            \par}

            Now it holds that $\B{a}\B{b}d$, but it does not hold that $\B{b}\B{a}\B{b}d$. So, general $\mathbf{a}$ would need to send the messenger once again to general $\mathbf{b}$. However, it can be intuitively seen that, regardless of how many messages the generals exchange, they will never be sure that the other will attack at dawn. This will be stated formally in Section \ref{sec:plan_ex}.
        \end{example}

    \subsection{The logic S5$_n$}\label{sec:s5}
        In the epistemic logic literature there exist many different axiomatizations of the concept of \emph{knowledge}. In this paper, we adopt the multimodal logic S5$_n$. Its axioms are\footnote{Even though \axiom{K}, \axiom{T} and \axiom{5} are sufficient to characterize S5$_n$, we include axiom \axiom{4} as it constitutes an important epistemic principle.}:
        
        {\centering
        $
            \begin{array}{l@{}l@{}r@{}}
                \axiom{K}\;\; & \B{i} (\varphi \rightarrow \psi) \rightarrow
                                (\B{i} \varphi \rightarrow \B{i} \psi)  			& \textnormal{(Distribution)}           \\
                \axiom{T} & \B{i} \varphi \rightarrow \varphi                       & \textnormal{(Knowledge)}              \\
                \axiom{4} & \B{i} \varphi \rightarrow \B{i} \B{i} \varphi           & \textnormal{(Positive introspection)} \\
                \axiom{5} & \neg \B{i} \varphi \rightarrow \B{i} \neg \B{i} \varphi & \textnormal{(Negative introspection)}
            \end{array}
        $
        \par}

        \noindent Axioms \axiom{T}, \axiom{4} and \axiom{5} correspond, to the following \emph{frame properties}: reflexivity ($ \forall u (u R_i u) $), transitivity ($ \forall u, v, w (u R_i v \wedge v R_i w \rightarrow u R_i w) $) and Euclidicity ($ \forall u, v, w (u R_i v \wedge u R_i w \rightarrow v R_i w) $). 
        Moreover, axioms \axiom{T} and \axiom{5} together entail symmetry ($ \forall u, v (u R_i v \rightarrow v R_i u) $). 
        Thus, accessibility relations in S5$_n$ are equivalence relations.
        We refer to epistemic states (resp., epistemic models, actions, frames) satisfying the axioms of a logic $L$ as $L$-states (resp., $L$-models, $L$-actions, $L$-frames). In the rest of the paper, we assume that the accessibility relations of epistemic states and actions are equivalence relations.

\subsection{Plan Existence Problem}\label{sec:plan_ex}
	We now define our problem, adapting the formulation in~\cite{conf/ijcai/Aucher2013}. 

    \begin{definition}[Planning Task]
    \label{def:planning_task}
        An \emph{(epistemic) planning task} is a triple $ T = (s_0, \actionSet,$ $ \varphi_g) $, where $ s_0 $ is an initial epistemic state; $ \actionSet $ is a finite set of actions; $ \varphi_g \in \Lang{C} $ is a \emph{goal formula}.
    \end{definition}

    Given a logic $ L $, an \emph{$L$-planning task} $ (s_0, \actionSet, \varphi_g) $ is a planning task where $s_0$ is an $L$-state and each action in $\actionSet$ is an $L$-action. We denote the class of $L$-planning tasks with $\mathcal{T}_L$.
    We remark that, given a generic logic $L$, the product update of an L-state with an L-action, in general, is not necessarily an L-state. This is not a desired outcome in general, since axioms model some principles of knowledge/belief that always need to be satisfied.
    For instance, this is the case of the logic KD45$_n$, that captures the concept of \emph{belief}. In the literature, there exist different approaches to guarantee the preservation of the KD45$_n$ frame properties after the product update. For instance, some techniques involve belief revision techniques \cite{workshop/nrac/Herzig2005}, whereas others focus on defining some additional conditions to impose to both states and actions \cite{conf/aaai/Son2015}.

    The case of our logic C-S5$_n$ is similar to that of KD45$_n$. In fact, the frame property corresponding to axiom \axiom{C} (see Equation \ref{eq:A-frame} in Section \ref{sec:new_logic}) is not guaranteed to hold after the application of an action. In this paper, rather than devising some technique to guarantee the preservation of frame property \ref{eq:A-frame}, we instead opt for a rollback-style approach: an action is not to be applied in a state if it would lead to violate the axioms of C-S5$_n$.
    This leads to the following definition.

    \begin{definition}[Solution]\label{def:solution}
        A \emph{solution} to an $L$-planning task $(s_0, \actionSet,$ $ \varphi_g)$ is a finite sequence $ \alpha_1, \dots, \alpha_m $ of actions of $\actionSet$ such that:
        \begin{compactenum}
            \item $ s_0 \otimes \alpha_1 \otimes \dots \otimes \alpha_m \models \varphi_g $, and
            \item For each $ 1 \leq k \leq m $, $ \alpha_k $ is applicable in $ s_0 \otimes \alpha_1 \otimes \dots \otimes \alpha_{k-1} $ and $s_0 \otimes \alpha_1 \otimes \dots \otimes \alpha_k$ is an $L$-state.
        \end{compactenum}
    \end{definition}
    
    \begin{definition}[Plan Existence Problem]\label{def:plan_ex_problem}
        Let $n \geq 1$ and $\mathcal{T}_L$ be a class of epistemic planning tasks for a logic $L$. \planex{$\mathcal{T}_L$}{$n$} is the following decision problem: ``Given an $L$-planning task $ T = (s_0, \actionSet, \varphi_g) \in \mathcal{T}_L $, where $|\agentSet|=n$, does $ T $ have a solution?''
    \end{definition}

    \begin{example}\label{ex:task}
        In Example~\ref{ex:update} we have seen that, intuitively speaking, the two generals can not coordinate a winning attack. We now state this formally. Let $T_\textnormal{coord} = (s_0, \actionSet, \varphi_g) $ be an S5$_n$-planning task, where $\actionSet = \{\textnormal{send}_{ab}, \textnormal{send}_{ba}\}$ and $\varphi = \CK{\{a,b\}}d$. Then, $T_\textnormal{coord}$ has no solution. In fact, for any number $k \geq 0$ of delivered messages, one can show by induction the following (where $h\geq 0$):
        \begin{compactitem}
            \item $k{=}2h$: $(\B{a}\B{b})^h\B{a} d $ holds in $s_k$, but not $(\B{b}\B{a})^{h+1}d$;
            \item $k{=}2h{+}1$: $(\B{b}\B{a})^{h+1}d $ holds in $s_k$, but not $(\B{a}\B{b})^{h+1}\B{a}d$.
        \end{compactitem}
        \noindent Thus, common knowledge can not be achieved between the two generals in a finite number of steps. However, any search algorithm would never terminate, since at each step there is exactly one applicable action that, when applied, results in a new S5$_n$-state.
    \end{example}

    \section{Semantic Approach and Commutativity}\label{sec:new_logic}
    In this section, we discuss in more detail the semantic approach and we show how it can be used to obtain decidability results. Then, we introduce and analyze the commutativity axiom in the context of epistemic planning.

    With the semantic approach, we aim at devising a new way to approach decidability in epistemic planning, which deviates from the common line of research in the literature focused on limiting the action theory syntactically (\eg by imposing a limit on the maximum modal depth of formulae). Our approach is motivated by the fact that to obtain decidable fragments of the general problem, one must appeal to strong syntactical constraints. To substantiate this claim, recall that the problem is still undecidable when the maximum modal depth allowed is set to $d{=}2$ (see Table \ref{tab:complexity1} for more details, where $\mathcal{T}(\ell, m)$ denotes the class of epistemic planning tasks where preconditions and postconditions have modal depth at most $ \ell$ and $ m$, respectively). This is clearly a strong limitation of syntactic approaches, as the isolated classes are too strong in many practical cases, where reasoning about the knowledge of others is required. Instead, the semantic approach does not limit the structure of formulae of the action theory, but rather relies on devising a suitable set of axioms that guarantee desirable properties on the structure of epistemic states (\eg bounded number of possible worlds). Since, in principle, there are many ways one can obtain such desirable properties by means of modal axioms, we argue that the semantic approach constitutes a fruitful avenue of research that can be further explored in many different ways.

    Towards this goal, we analyze in detail the commutativity axiom. The key insight behind the definition of such axiom is that, in the logic S5$_n$, there is no rule or principle that describes how the knowledge of one agent should interact with the knowledge of another agent. Hence there is no restriction on the ability of agents to reason about the higher-order knowledge they possess about the knowledge of others. 
    This is clear in Example \ref{ex:update}, where at each step $k$ we obtain an epistemic state $s_k$ that contains a chain of worlds of the form $ w_1 R_{i} w_2 R_{j} w_3 R_{i} \dots w_k $ (for $i,j \in \{\mathsf{a}, \mathsf{b}\}, i \neq j $), that intuitively represents $i$'s perspective about $j$'s perspective about $i$'s perspective, and so forth. This idea has been exploited for building undecidability proofs of the plan existence problem in the logic S5$_n$, by showing a reduction from the halting problem of Turing machines \cite{journals/jancl/Bolander2011} and Minsky two-counter machines \cite{conf/ijcai/Aucher2013}.

	To weaken this reasoning power, we introduce a principle that governs the capability of agents to reason about the knowledge of others, which is captured by the following interaction axiom (where $i{\neq}j$):
    \begin{equation*}
        \begin{array}{lll}
            \axiom{C} & \B{i} \B{j} \varphi \rightarrow \B{j} \B{i} \varphi & \textnormal{(Commutativity)}
        \end{array}
    \end{equation*}

    \noindent The commutativity axiom is well-known in many-dimensional modal logics where it is part of the axiomatisation of the product of two modal logics \cite{book/nhpc/Gabbay2003}.
    Here, we adopt it with a novel epistemic connotation. Namely, we can read \axiom{C} as follows: whenever an agent $i$ knows that another agent $j$ knows that $\varphi$, then $j$ knows that $i$ knows \emph{too} that $\varphi$.
    Thus, intuitively, axiom \axiom{C} defines a principle of \emph{commutativity} in the knowledge that agents have about the knowledge of others. 
    This intuition is formalized and proved in the next section (see Lemma~\ref{lem:ck-n} and Theorem~\ref{th:ck}).
    
    This axiom is instrumental in proving decidability of the plan existence problem. Moreover, it provides a useful principle for two main reasons. 
    First, as we mentioned above, this axiom allows to govern the reasoning power of agents. As it turns out, in this way we obtain a \emph{finitary non-fixpoint} characterization of common knowledge (see Theorem \ref{th:ck}), which concretely shows the power of knowledge commutativity.
    Second, this principle constitutes a reasonable assumption in several \emph{cooperative multi-agent planning tasks} \cite{journals/csur/Torreno2017} where agents are able to communicate or monitor each other. In fact, when autonomous agents cooperate to reach a shared goal, then they are expected to behave in such a way that the effects of their actions are observable by others. In other words, acting in a cooperative context results into a transparent behaviour of agents, which in turn well fits with the concept of knowledge commutativity.

    We call C-S5$_n$ the logic S5$_n$ augmented with axiom \axiom{C}.
    As a final remark, notice that axiom \axiom{C} is a Sahlqvist formula and it corresponds to the following frame property:
    \begin{equation}\label{eq:A-frame}
        \forall u, v, w (u R_j v \wedge v R_i w \rightarrow \exists x (u R_i x \wedge x R_j w))
    \end{equation}
    
    \begin{proposition}
        The logic C-S5$_n$ is sound and complete with the class of reflexive, symmetric and transitive epistemic models that enjoy property (\ref{eq:A-frame}).
    \end{proposition}

    \section{Epistemic Planning with Commutativity}\label{sec:decidability}
    This section is organized as follows. First, we prove the decidability of the plan existence problem under the logic C-S5$_n$. Then, in Section \ref{sec:general-comm}, we consider two generalizations of the commutativity axiom and we analyze their impact in the decidability of the plan existence problem. We first recall the following result:

    \begin{thm}[\cite{conf/ijcai/Aucher2013}]\label{th:aucher-bolander}
        For any $n > 1$, \planex{$\mathcal{T}_{\textnormal{S5}}$}{$n$} is undecidable.
    \end{thm}

    We now focus on the logic C-S5$_n$, assuming that $n>1$. We begin by giving some preliminary results.

    \begin{lemma}\label{lem:slide-box}
        Let $ G = \{i_1, \dots, i_m\} \subseteq \agentSet $, with $ m \geq 2 $ and let $ \vec{v} \in G^* $ ($ |\vec{v}| = \lambda  \geq 2$). Let $ \pi $ and $ \rho $ be two permutations of elements of $ \vec{v} $. Then, for any $\varphi$, in the logic C-S5$_n$ the following is a theorem:
        \begin{equation*}
            \B{\pi_1} \dots \B{\pi_\lambda} \varphi \leftrightarrow \B{\rho_1} \dots \B{\rho_\lambda} \varphi
        \end{equation*}
    \end{lemma}

    For a group $G$ of agents, the knowledge of agent $i_1$ about what agent $i_2$ knows about what agent $i_3$ knows, and so forth up to agent $i_m$, \emph{coincides} exactly with the knowledge of any of the agents of $G$ about what some other agent knows about some third agent, and so forth up to the $m$-th agent. That is, higher-order knowledge involving a group of agents is independent from the order in which we consider said agents (\ie it commutes).
    In other words, Lemma \ref{lem:slide-box} shows us that we can \emph{rearrange} the order of any sequence of boxes in a formula and obtain an equivalent one.

    \begin{lemma}\label{lem:ck-n}
        Let $ G = \{i_1, \dots, i_m\} \subseteq \agentSet $, with $ m \geq 2 $. In the logic C-S5$_n$, for any $\varphi$ and $ \vec{v} \in G^* $ we have that $ \B{i_1} \dots \B{i_m} \varphi \rightarrow \B{v_1} \cdots \B{v_{|\vec{v}|}} \varphi $ is a theorem.
    \end{lemma}

	Lemma~\ref{lem:ck-n} provides the basis to show a first important result related to common knowledge in C-S5$_n$. Namely, we obtain a \emph{finitary non-fixpoint} characterization of common knowledge.

    \begin{thm}\label{th:ck}
        Let $ G = \{i_1, \dots, i_m\} \subseteq \agentSet $, with $ m \geq 2 $. In the logic C-S5$_n$, for any $\varphi$, the formula $ \B{i_1} \dots \B{i_m} \varphi \leftrightarrow \CK{G} \varphi $ is a theorem.
        \begin{proof} 
            ($\Leftarrow$) This follows by definition of common knowledge; ($\Rightarrow$) this immediately follows by Lemma \ref{lem:ck-n}.
        \end{proof}
    \end{thm}

    \begin{corollary}\label{cor:diameter}
        Let $ G = \{i_1, \dots, i_m\} \subseteq \agentSet $, with $ m \geq 2 $. 
        In an C-S5$_n$-model, for any $\vec{v} \in G^*$, we have that if $w R_{v_1} \circ \ldots \circ R_{v_{|\vec{v}|}} w'$, then $w R_{i_1} \circ \dots \circ R_{i_m} w'$.
    \end{corollary}
    
    The statement above directly follows from the contrapositive of the implication in Lemma \ref{lem:ck-n}, under the assumption of minimality of states (w.r.t. bisimulation).
    Intuitively, this states that in a C-S5$_n$-model, given any subset of $m\geq 2$ agents, if a world is reachable in an arbitrary number of steps, then it is also reachable in exactly $m$ steps. Thus, in general, any pair of worlds of a C-S5$_n$-model that are reachable from one another are connected by a path of length \emph{at most $n$}. This property suggests the existence of some boundedness property on the size of C-S5$_n$-states. Indeed, we exploit Corollary \ref{cor:diameter} to prove the following lemma.

    \begin{lemma}\label{lem:bounded-bisim}
        Let $(M, W_d)$ be an C-S5$_n$-state, with $M=(W,R,V)$. For any $w,v \in W$, we have that $ w \bisim_{n+1} v \Leftrightarrow w \bisim v$.
    \end{lemma}

    Lemma \ref{lem:bounded-bisim} shows that, in the logic C-S5$_n$, to verify whether two worlds are bisimilar, we only need to check their neighborhoods up to distance $n+1$. We exploit this intuition, under the assumption of bisimulation minimality, to prove the following lemma.

    \begin{lemma}\label{lem:char-formulae}
        Let $(M,W_d)$ be a bisimulation-contracted C-S5$_n$-state, with $M=(W,R,V)$. Then, $|W|$ is bounded in $n$ and $|\atomSet|$.
    \end{lemma}

    \noindent Having a bound in the number of worlds of a C-S5$_n$-state immediately provides us with the following decidability result.

    \begin{thm}\label{th:dec}
    	For any $n{>}1$, \planex{$\mathcal{T}_{\textnormal{C-S5}}$}{$n$} is \emph{decidable}.
        \begin{proof}
            Let $T \in \mathcal{T}_{\textnormal{C-S5}_n}$ be an epistemic planning task. By Lemma \ref{lem:char-formulae}, it follows that we can perform a breadth-first search on the search space that would only visit a finite number of epistemic states (up to bisimulation contraction) to find a solution for $T$. Thus, we obtain the claim.
        \end{proof}
    \end{thm}

    The following example shows that in C-S5$_n$ we can effectively obtain an answer to the Coordinated Attack Problem, hence showing that common knowledge can not be achieved by the two generals.

    \begin{example}\label{rem:1}
        As shown in Example \ref{ex:task}, the S5$_n$-planning task $T_\textnormal{coord}$ has no solution, but any search algorithm would never terminate. We now consider the C-S5$_n$-planning task $T^A_\textnormal{coord} = (s_0, \actionSet, \varphi_g)$, with $s_0$, $\actionSet$ and $\varphi_g$ defined as in Example \ref{ex:task} (notice that $s_0$ is an C-S5$_n$-state and that send$_{ab}$ and send$_{ba}$ are both C-S5$_n$-actions). We immediately note that the epistemic state $s_1$ of Example \ref{ex:update} is \emph{not} an C-S5$_n$-state. Thus, by Definition \ref{def:solution} and since send$_{ab}$ is the only applicable action in $s_0$, any search algorithm would immediately stop returning the answer ``\emph{no}''. Hence, in the logic C-S5$_n$, one can conclude that it is impossible for the two generals to coordinate an attack in a finite number of steps.
    \end{example}
    
    \subsection{Generalizing the Principle of Commutativity}\label{sec:general-comm}
    Although the commutativity axiom is better fitting for tight-knit groups of agents, it may be less suited for representing more loosely organized groups.
    Thus, having established that adding axiom \axiom{C} to S5$_n$ leads to decidability of the plan existence problem, we investigate two generalized principles of commutativity, namely \emph{$b$-commutativity} and \emph{weak commutativity}. In what follows, we consider such generalizations and we provide (un)decidability results of their corresponding plan existence problems.
        
        \paragraph{$b$-Commutativity}\label{par:b-comm}
        Let $b{>}1$ be a fixed constant. Then, we define the following axiom:
        \begin{equation*}
            \begin{array}{lll}
                \axiom{C$^b$} & (\B{i} \B{j})^b \varphi \rightarrow (\B{j} \B{i})^b \varphi & \textnormal{($b$-Commutativity)}
            \end{array}
        \end{equation*}

        \noindent We call C$^b$-S5$_n$ the logic S5$_n$ augmented with axiom \axiom{C$^b$}. Axiom \axiom{C$^b$} generalizes commutativity by considering an arbitrary fixed amount of repetitions of box operators. Indeed, notice that \axiom{C$^1$} $=$ \axiom{C}.
        Moreover, since every $\Box_i$ is a monotone modality (\ie from $\varphi \rightarrow \psi$ we can infer $\B{i}\varphi \rightarrow \B{i}\psi$), it is easy to see that each axiom \axiom{C$^{b+1}$} leads to a weaker logic than axiom \axiom{C$^b$}, and that every logic C$^b$-S5$_n$ is weaker than C-S5$_n$. 
        One could hope that the plan existence problem remains decidable when replacing axiom \axiom{C} with \axiom{C$^b$}. But this is not true in general. In fact, we prove that it remains decidable for $n{=}2$ and any $b{>}1$ (Theorem \ref{th:dec-b-2}) and that it becomes undecidable for any $n{>}2$ and $b{>}1$ (Theorem \ref{th:undec-b-n}).
        Due to space constraints, we only provide the proof sketches (full proofs are available in the arXiv Appendix).

        \begin{thm}\label{th:dec-b-2}
            For any $b{>}1$, \planex{$\mathcal{T}_{\textnormal{C$^b$-S5}}$}{$2$} is \emph{decidable}.
            
            \begin{proof}
                \emph{(Sketch)} Analogous to the case of the logic C-S5$_n$. Namely, we can prove the correspondent versions of Lemma 
                \ref{lem:ck-n}, Theorem \ref{th:ck}, Corollary \ref{cor:diameter} and Lemmata \ref{lem:bounded-bisim} and \ref{lem:char-formulae}. The claim follows by combining these results as in Theorem \ref{th:dec}.
            \end{proof}
        \end{thm}

        \begin{example}
            As the Coordinated Attack Problem involves exactly two agents, for any $b{>}1$, we can define the C$^b$-S5$_2$-planning task $T^{\textnormal{C}^b}_\textnormal{coord} = (s_0, \actionSet, \varphi_g)$, with $s_0$, $\actionSet$ and $\varphi_g$ defined as in Example \ref{rem:1}. Then, as above, we note that the epistemic state $s_{1+2(b-1)}$ of Example \ref{ex:task} is \emph{not} a C$^b$-S5$_2$-state. Thus, by Definition \ref{def:solution} and since send$_{ab}$ is the only applicable action in $s_{2(b-1)}$, a search algorithm would return the answer ``\emph{no}'' in exactly $2(b{-}1)$ steps.
        \end{example}

        \begin{thm}\label{th:undec-b-n}
            For any $n{>}2$, $b{>}1$, \planex{$\mathcal{T}_{\textnormal{C$^b$-S5}}$}{$n$} is \emph{undecidable}.

            \begin{proof}
                \emph{(Sketch)} We adapt the proof in \cite[Section~6]{conf/ijcai/Aucher2013}, of the undecidability of epistemic planning in the logic $S5_n$ ($n > 1$). It is an elegant reduction from the halting problem of Minsky two-counter machines \cite{book/ph/Minsky1967} to the plan existence problem.
                
                We prove our result for the logic C$^2$-S5$_3$ (\ie having $b=2$ and $n=3$). Since C$^2$-S5$_3$-models are also C$^b$-S5$_n$-models for any $n > 3$ and $b > 2$, our results hold for any combination of the values of $n \geq 3$ and $b \geq 2$. Given a two-counter machine $M$, the procedure follows three steps:
                \begin{compactenum}
                    \item We define an encoding for integers and configurations;
                    \item We build a finite set of actions for encoding the computation function $f_M$; and
                    \item We combine the previous steps and we encode the halting problem as an C$^2$-S5$_3$-planning task.
                \end{compactenum}
                Finally, the claim follows by showing that the resulting planning task has a solution iff $M$ halts.
            \end{proof}
        \end{thm}

        These results show that it is not straightforward to generalize knowledge commutativity and to maintain the decidability of the plan existence problem. However, Theorem \ref{th:dec-b-2} suggests that the logic C$^b$-S5$_n$ could result into interesting developments in contexts where only two agents are involved (\eg epistemic games).

        \paragraph{Weak commutativity}\label{par:weak-comm}
        Let $1{<}\ell{\leq} n$ be a fixed constant. Let $\langle i_1, \dots, i_\ell \rangle$ be a sequence of agents with no repetitions, and let $\pi$ be any permutation of this sequence. Then, we define the following axiom (for any such $\pi$):
        \begin{equation*}
            \begin{array}{lll}
                \axiom{wC$_\ell$} & \B{i_1} \dots \B{i_\ell}\varphi \rightarrow \B{\pi_{i_1}} \dots \B{\pi_{i_\ell}}\varphi & \textnormal{(Weak comm.)}
            \end{array}
        \end{equation*}

        \noindent We call wC$_\ell$-S5$_n$ the logic S5$_n$ augmented with axiom \axiom{wC$_\ell$}. Axiom \axiom{wC$_\ell$} generalizes commutativity by extending it to more than two agents, whereas \axiom{C} corresponds to taking $\ell=2$. 
        Indeed, notice that \axiom{wC$_2$} $=$ \axiom{C}. Moreover, since every $\Box_i$ is a monotone modality, it is easy to see that each axiom \axiom{wC$_{\ell+1}$} leads to a weaker logic than axiom \axiom{wC$_\ell$}, and that every logic wC$_\ell$-S5$_n$ is weaker than wC-S5$_n$.

        By considering this form of generalization of axiom \axiom{C}, we are able to provide a decidability result that holds for any $1 < \ell \leq n$. The arguments adopted by the proof are similar to those of Theorem \ref{th:dec}.

        \begin{thm}\label{th:dec-l}
            For any $n{>}1$ and $1 {<} \ell {\leq} n$, \planex{$\mathcal{T}_{\textnormal{wC$_\ell$-S5}}$}{$n$} is \emph{decidable}.
            \begin{proof}
                \emph{(Sketch)} As in the proof of Theorem \ref{th:dec}, we can prove the correspondent versions of Lemmata \ref{lem:slide-box}, \ref{lem:ck-n}, Theorem \ref{th:ck} and Corollary \ref{cor:diameter}. From these results, we obtain that any pair of worlds of a wC$_\ell$-S5$_n$-model that are reachable from one another are connected by a path of length \emph{at most $n$}. Hence, we show that Lemmata \ref{lem:bounded-bisim} and \ref{lem:char-formulae} hold also in the logic wC$_\ell$-S5$_n$ (for any $\ell > 1$). Thus, to obtain the claim, we use Lemmata \ref{lem:bounded-bisim} and \ref{lem:char-formulae} by combining them as in Theorem \ref{th:dec}.
            \end{proof}
        \end{thm}

        To summarize, $b$-commutativity and weak commutativity constitute two generalizations of axiom \axiom{C}. All of the above decidability results are outlined in Table \ref{tab:complexity2}.

    \section{Epistemic Planning Systems}\label{sec:systems}
    In this section, we look at two well-known epistemic planning systems that adopt the DEL semantics and we show their decidability by applying our previous results. These are $m\mathcal{A}^*$ \cite{journals/corr/Baral2015} and the framework by Kominis and Geffner \cite{conf/aips/Kominis2015}.
    While decidability is already known for the latter, the decidability of the former is still an open problem. In what follows, we show that both these systems are captured by our setting when knowledge is considered, namely under S5$_n$ axioms. We begin by briefly introducing the two systems.
        
    \begin{figure}[t]
        \centering{
        \begin{tikzpicture}
            \input{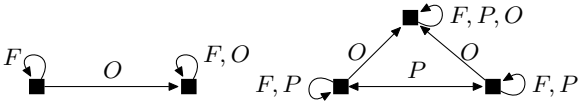}
        \end{tikzpicture}
        }
        \caption{Frames of $m\mathcal{A}^*$ event models for ontic actions (left) and sensing/announcement actions (right).}
        \label{fig:ma_star}
    \end{figure}

    \noindent
    \textbf{1.} The epistemic planning framework $m\mathcal{A}^*$ \cite{journals/corr/Baral2015} features three action types: 
    \emph{ontic}, \emph{sensing} and \emph{announcement} actions. 
    In $m\mathcal{A}^*$, agents are partitioned in three sets: \emph{fully observant} agents ($F$), that are able to observe the action corresponding to an event, \emph{partially observant} agents ($P$) that only know about the execution of an action, but not the effects, and \emph{oblivious} agents ($O$), that are ignorant about the fact that the action is taking place. 
    When oblivious agents are considered, however, event models fall beyond C-S5$_n$ (they have KD45$_n$-frames), as they are not \emph{symmetric} (see Figure \ref{fig:ma_star}).
    Consequently, we restrict ourselves to a fragment of $m\mathcal{A}^*$ that includes \emph{public} ontic actions and \emph{semi-private} sensing and announcement actions. This is achieved by removing from the event models of Figure \ref{fig:ma_star} all events considered possible by oblivious agents. It is easy to see that the frames of the resulting event models are indeed S5$_n$-frames.
    We call the resulting system the \emph{S5$_n$-fragment} of $m\mathcal{A}^*$, and we denote with $\mathcal{T}_{\textnormal{S5$_n$-}m\mathcal{A}^*}$ the class of planning tasks of such system.
    
    \noindent
    \textbf{2.} 
    Kominis and Geffner \cite{conf/aips/Kominis2015} describe a system for handling beliefs in multi-agent scenarios. They describe three types of actions: \emph{do}, \emph{update}, and \emph{sense}. Although their formulation is not given in terms of DEL semantics, the authors briefly describe the event models corresponding to each action type (Figure \ref{fig:kom15}). We denote with $\mathcal{T}_{\textnormal{KG}}$ the class of planning tasks of such system. Differently from $m\mathcal{A}^*$, all the described event models already have S5$_n$-frames.

    \begin{figure}[t]
        \centering{
            \begin{tikzpicture}
                \input{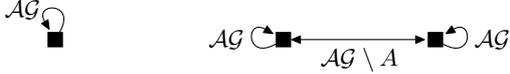}
            \end{tikzpicture}
        }
        \caption{Frames of event models in \cite{conf/aips/Kominis2015} for \emph{do} and \emph{update} actions (left) and \emph{sense} (right).}
        \label{fig:kom15}
    \end{figure}

    We show that the two described systems fall within our logic, thus proving their decidability (see arXiv Appendix for full proofs):
    \begin{lemma}\label{lem:systems}
        $\mathcal{T}_{\textnormal{S5$_n$-}m\mathcal{A}^*} \subseteq \mathcal{T}_{\textnormal{C-S5}}$ and $\mathcal{T}_{\textnormal{KG}} \subseteq \mathcal{T}_{\textnormal{C-S5}}$.
    \end{lemma}

    \begin{corollary}
        For any $n>1$, \planex{$\mathcal{T}_{\textnormal{S5$_n$-}m\mathcal{A}^*}$}{$n$} and \planex{$\mathcal{T}_{\textnormal{\textbf{KG}}}$}{$n$} are \emph{decidable}.
    \end{corollary}

    \section{Related Works}\label{sec:rel_works}
    The DEL semantics has been widely exploited to obtain complexity and decidability results for the plan existence problem \cite{conf/ijcai/Aucher2013,journals/corr/Aucher2014,journals/jancl/Bolander2011,journals/ai/Bolander2020,conf/ijcai/Bolander2015,conf/ijcai/Charrier2016,conf/lori/Lowe2011,conf/ijcai/Yu2013}.
    We identify three main lines of research.
    The first one restricts the class of actions that are allowed in planning tasks by means of syntactical conditions. In particular, some approaches constrain actions to be \emph{purely epistemic} (\ie without postconditions), while others introduce limitations to the modal depth preconditions and/or postconditions. Following the notation of \cite{journals/ai/Bolander2020}, we denote with $\mathcal{T}(\ell, m)$ the class of epistemic planning tasks in which the modal depth of the preconditions are $\leq \ell$, and that of the postconditions are $\leq m$. Similarly, planning tasks with purely epistemic actions are denoted with $\mathcal{T}(\ell,-1)$. We report the main results in Table \ref{tab:complexity1}.

    \begin{table}[t]
        \centering
        \begin{tabular}{|l|l|}
            \hline
            \planex{$\mathcal{T}(0,-1)$}{$m$} & \textsc{pspace}-complete \cite{conf/ijcai/Charrier2016} \\ \hline
            \planex{$\mathcal{T}(1,-1)$}{$n$} & \textsc{unknown}         \cite{conf/ijcai/Charrier2016} \\ \hline
            \planex{$\mathcal{T}(2,-1)$}{$n$} & \textsc{undecidable}     \cite{conf/ijcai/Charrier2016} \\ \hline
            \planex{$\mathcal{T}(0,0)$}{$n$}  & \textsc{decidable}       \cite{conf/ijcai/Yu2013,journals/corr/Aucher2014} \\ \hline
            \planex{$\mathcal{T}(1,0)$}{$n$}  & \textsc{decidable}       \cite{journals/ai/Bolander2020} \\ \hline
        \end{tabular}
        \caption{Decidability and complexity results of plan existence problem based on the \emph{syntactical} approach.}
        \label{tab:complexity1}
    \end{table}

    The second line of research pivots both on syntactical limitations to formulae and on constraining the frames of event models (\eg singletons, chains, trees). In particular, \planex{$\mathcal{T}(0,-1)$}{$n$} is \textsc{np}-complete on singletons and chains and it is \textsc{pspace}-complete on trees \cite{conf/ijcai/Bolander2015}, whereas \planex{$\mathcal{T}(\ell, m)}{n}$ (for any $\ell, m \geq 0$) on singletons is \textsc{pspace}-hard \cite{phd/dtu/Jensen2014}.

    Finally, the third line of research revolves around the choice of the logic for epistemic models and actions \cite{conf/ijcai/Aucher2013}. This approach is more similar to the one we adopted in this paper, with the difference that the logics considered in previous works are a combination of standard and well-known axioms of epistemic logic (see Section \ref{sec:new_logic}). We report the main results in Table \ref{tab:complexity2}. For a more detailed analysis of complexity and decidability results in epistemic planning we refer the interested reader to \cite{journals/ai/Bolander2020}.

    \begin{table}[t]
        \centering
        \begin{tabular}{|c|c|}
            \hline
            Logic                                                 & Decidability                                             \\ \hline
            K$_n$, K$_n$, KT$_n$, K4$_n$, K45$_n$, S4$_n$, S5$_n$ & \textsc{undecidable} \cite{conf/ijcai/Aucher2013}        \\ \cline{1-1}
            C$^b$-S5$_n$ ($n{>}2$)                                & \cellcolor{gray!35} \textsc{undecidable}                 \\
            C$^b$-S5$_2$                                          & \cellcolor{gray!15}                                      \\ \cline{1-1}
            wC$_\ell$-S5$_n$                                      & \cellcolor{gray!15}                                      \\ \cline{1-1}
            C-S5$_n$                                              & \cellcolor{gray!15} \multirow{-3}{*}{\textsc{decidable}} \\ \hline
        \end{tabular}
        \caption{Decidability results of plan existence problem based on the semantic approach, compared to our results (in gray).}
        \label{tab:complexity2}
    \end{table}

    \section{Conclusions}
    The paper presents novel decidability results in epistemic planning. The approach adopted in this work deviates from previous ones, where syntactical conditions are imposed to actions. In particular, we pursue a novel semantic approach by introducing a principle of knowledge commutativity that is well suited for cooperative multi-agent planning contexts. In this way, we govern the extent to which agents can reason about the knowledge of their peers. This results in a boundedness property of the size of epistemic states, which in turn guarantees that the search space is finite.
    Starting from this key result, we studied decidability of the plan existence problem under different generalizations of the commutativity axiom, showing both positive and negative results.

    Notably, our decidability results are orthogonal to the problem of preservation of commutativity during planning. In this paper we adopt the baseline strategy of rejecting action sequences that visit invalid states. A natural follow up is then to incorporate more sophisticated techniques to revise/repair such invalid states, in the style of belief revision for KD45$_n$ \cite{workshop/nrac/Herzig2005} and to single out fragments of C-S5$_n$ where preservation is guaranteed by design \cite{conf/aaai/Son2015}.

    In the future, we plan on further investigating our semantic approach by formulating other properties to add to the logic of knowledge. In particular, we are interested into defining new generalizations of the commutativity axiom to obtain broader fragments that maintain decidability of the plan existence problem. Moreover, we are interested into verifying if and how this approach can be suitably recast in a doxastic setting, where knowledge is replaced by \emph{belief}. This is not a trivial task, as the results of this paper do not readily apply to the logic KD45$_n$. We also intend to delve into a fine-grained analysis of the computational complexity of $\mathcal{T}_{\textnormal{C-S5}_n}$ and to compare it with the current results in the literature.


    \ack This research has been partially supported by the Italian Ministry of University and Research (MUR) under PRIN project PINPOINT Prot. 2020FNEB27, and by the Free University of Bozen-Bolzano with the ADAPTERS project.

    \newpage












\newtheorem{apptheorem}{Theorem}[section]
\newtheorem{applemma}{Lemma}[section]
\newtheorem{appcorollary}{Corollary}[section]
\newtheorem{appdefinition}{Definition}[section]
\newtheorem*{appclaim*}{Claim}

\newcounter{mainlemmacount}
\newcounter{maintheoremcount}
\newcounter{maincorollarycount}

\newcommand{\setlemmacountertoref}[1]{%
    \setcounterref{mainlemmacount}{#1}%
    \setcounter{lemma}{\themainlemmacount}%
    \addtocounter{lemma}{-1}%
}

\newcommand{\settheoremcountertoref}[1]{%
    \setcounterref{maintheoremcount}{#1}%
    \setcounter{thm}{\themaintheoremcount}%
    \addtocounter{thm}{-1}%
}

\newcommand{\setcorollarycountertoref}[1]{%
    \setcounterref{maincorollarycount}{#1}%
    \setcounter{corollary}{\themaincorollarycount}%
    \addtocounter{corollary}{-1}%
}

\externaldocument[main:]{../main}




    \appendix
    \section*{Appendix}
    \noindent In what follows, we provide the full proofs of our results. The sections of the Appendix are named after the corresponding (sub)sections of the paper. To enhance clarity, results that are \emph{not} present in the paper are numbered referring to the correspondent Appendix section.

    \section{\nameref*{sec:decidability}}

    \setlemmacountertoref{lem:slide-box}
    \begin{lemma}\label{lem:slide-box-proof}
        Let $ G = \{i_1, \dots, i_m\} \subseteq \agentSet $, with $ m \geq 2 $ and let $ \vec{v} \in G^* $ ($ |\vec{v}| = \lambda \geq 2$). Let $ \pi $ and $ \rho $ be two permutations of elements of $ \vec{v} $. Then, for any $\varphi$, in the logic C-S5$_n$ the following is a theorem:
        \begin{equation*}
            \B{\pi_1} \dots \B{\pi_\lambda} \varphi \leftrightarrow \B{\rho_1} \dots \B{\rho_\lambda} \varphi
        \end{equation*}

        \begin{proof}
            First, we notice that in the logic C-S5$_n$, for any formula $\varphi$, the following formula is a theorem:
            \begin{equation}\label{eq:comm-2}
                \B{i} \B{j} \varphi \leftrightarrow \B{j} \B{i} \varphi
            \end{equation}

            \noindent This immediately follows from axiom \axiom{C}.

            Second, by construction, we have that for each $ \pi_i $ there exists $ \rho_{k_i} $ such that $ \rho_{k_i} = \pi_i $. Consider $ \rho_{k_1} = \pi_1 $. Then, by iterating Equation \ref{eq:comm-2}, we obtain:
            \begin{align*}
                                &~ \B{\rho_1} \dots \B{\rho_{{k_1}-1}} \B{\rho_{k_1}} \B{\rho_{{k_1}+1}} \dots \B{\rho_\lambda} \varphi \\
                \leftrightarrow &~ \B{\rho_1} \dots \B{\rho_{k_1}} \B{\rho_{{k_1}-1}} \B{\rho_{{k_1}+1}} \dots \B{\rho_\lambda} \varphi \\
                \dots           &~                                                                                                \\
                \leftrightarrow &~ \B{\rho_1} \B{\rho_{k_1}} \dots \B{\rho_{{k_1}-1}} \B{\rho_{{k_1}+1}} \dots \B{\rho_\lambda} \varphi \\
                \leftrightarrow &~ \B{\rho_{k_1}} \B{\rho_1} \dots \B{\rho_{{k_1}-1}} \B{\rho_{{k_1}+1}} \dots \B{\rho_\lambda} \varphi
            \end{align*}
            By repeating this manipulation for $ \pi_2, \dots \pi_m $, we obtain the conclusion.
        \end{proof}
    \end{lemma}

    \setlemmacountertoref{lem:ck-n}
    \begin{lemma}\label{lem:ck-n-proof}
        Let $ G = \{i_1, \dots, i_m\} \subseteq \agentSet $, with $ m \geq 2 $. In the logic C-S5$_n$, for any $\varphi$ and $ \vec{v} \in G^* $ we have that $ \B{i_1} \dots \B{i_m} \varphi \rightarrow \B{v_1} \cdots \B{v_{|\vec{v}|}} \varphi $ is a theorem.

        \begin{proof}
            The proof is by induction on $|\vec{v}|$.
            For the base case ($|\vec{v}|=0$) we have that the formulae $ \B{i_h} \B{i_{h+1}} \dots \B{i_m} \varphi \rightarrow \B{i_{h+1}} \dots \B{i_m} \varphi $ ($ 1 \leq h < m $) and $ \B{i_m} \varphi \rightarrow \varphi $ are instances of \axiom{T}. Together with propositional reasoning, we get that $ \B{i_1} \dots \B{i_m} \varphi \rightarrow \varphi $ is a theorem.

            Let now $|\vec{v}| = \lambda$ and suppose, by inductive hypothesis, that $\B{i_1} \dots \B{i_m} \varphi \rightarrow \B{v_1} \cdots \B{v_{\lambda}} \varphi$ is a theorem (for any formula $\varphi$). We now show that, for each $ j \in G $, the formula $\B{i_1} \dots \B{i_m} \varphi \rightarrow \B{v_1} \cdots \B{v_{\lambda}} \B{j} \varphi$ is also a theorem. By inductive hypothesis, substituting $\varphi$ with $\B{j} \varphi$, the following is a theorem:
            \begin{equation*}
                \B{i_1} \dots \B{i_m} \B{j} \varphi \rightarrow \B{v_1} \cdots \B{v_{\lambda}} \B{j} \varphi.
            \end{equation*}

            \noindent Since $ j {\in} G $, there exists $ h {\in} \{1, \dots m\} $ such that $ j = i_h $. From this and Lemma \ref{lem:slide-box-proof}, we can rewrite the antecedent of the above implication, $ \B{i_1} \dots \B{i_m} \B{j} \varphi $, as $ \B{i_1} \dots \B{j} \B{j} \dots \B{i_m} \varphi $. Moreover, it is easy to prove that from axioms \axiom{T} and \axiom{4} the following formula is a theorem (for any formula $\varphi$):
            \begin{equation}\label{eq:box-absoption}
                \B{j} \varphi \leftrightarrow \B{j} \B{j} \varphi.
            \end{equation}
            Thus, we can rewrite the above formula as $ \B{i_1} \dots \B{j} \dots \B{i_m} \varphi $, which is simply $ \B{i_1} \dots \B{i_m} \varphi $. Finally, we obtain that the following is a theorem:
            \begin{equation*}
                \B{i_1} \dots \B{i_m} \varphi \rightarrow \B{v_1} \cdots \B{v_{\lambda}} \B{j} \varphi.
            \end{equation*}
            This is the required result.
        \end{proof}
    \end{lemma}

    \setlemmacountertoref{lem:bounded-bisim}
    \begin{lemma}\label{lem:bounded-bisim-proof}
        Let $(M, W_d)$ be an C-S5$_n$-state, with $M=(W,R,V)$. For any $w,v \in W$, we have that $ w \bisim_{n+1} v \Leftrightarrow w \bisim v$.

        \begin{proof}
                        
            Clearly, if $ w \bisim v$, then $w \bisim_{n+1} v$. 
            For the other direction, assume $w \bisim_{n+1} v$. First recall that there exists a path between any two worlds $w$ and $v$ of length at most $n$ (Corollary \ref{cor:diameter}). We refer to this property as ($\dagger$). By contradiction, assume that that it is not the case that $ w \bisim v $, namely there exist two worlds $w',v' \in W$ such that:
            \begin{itemize}
                \item $w' \bisim_0 v'$;
                \item $w'$ is reached by a path $\pi_w$ starting from $w$ (with $|\pi_w| {=} \ell$);
                \item $v'$ is reached by a path $\pi_v$ starting from $v$ (with $|\pi_v| {=} \ell$);
                \item There exists a world $w'' \in W$ such that $w' R_{i_{\ell+1}} w''$ (with ${i_{\ell+1}} \in \agentSet$), such that for all worlds $v'' \in W$ such that $v' R_{i_{\ell+1}} v''$, it is not the case that $w''\bisim_0 v''$ and, thus, that $w'\bisim_1 v'$ (or vice-versa, swapping $w'$ with $v'$ and $w''$ with $v''$).
            \end{itemize}
            %
            Let us denote these two paths as:
            $\pi_w = w R_{i_1} \circ \dots \circ R_{i_\ell} w' R_{i_{\ell+1}} w''$ and
            $\pi_v = v R_{i_1} \circ \dots \circ R_{i_\ell} v' R_{i_{\ell+1}} v''$, with each $i_x\in \agentSet$ for $1 \leq x \leq \ell+1$.

            Clearly, $\ell \geq n+1$ otherwise this would contradict the hypothesis that $ w \bisim_{n+1} v $. 
            Assume $\ell=n+1$. Thus, $|\pi_w| = |\pi_v| = n+2$. 
            We now show that $w'\bisim_1 v'$. From ($\dagger$) it follows that there exist two shorter paths
            $\pi'_w = w R_{j_1} \circ \dots \circ R_{j_n} w' R_{j_{n+1}} w''$ and
            $\pi'_v = v R_{j_1} \circ \dots \circ R_{j_n} v' R_{j_{n+1}} v''$, with each $j_x\in \agentSet$ for $1 \leq x \leq n+1$.

            Since by hypothesis $w \bisim_{n+1} v$, this means that $v''$ as above exists and also that $w'' \bisim_0 v''$ for any such $w''$ and $v''$.
            This implies $w'\bisim_1 v'$ and thus $w\bisim_{n+2} v$.
            Since the same argument applies for any $\ell > n+1$, we obtain that $w \bisim v$. 
            
        \end{proof}
    \end{lemma}

    \setlemmacountertoref{lem:char-formulae}
    \begin{lemma}\label{lem:char-formulae-proof}
        Let $(M,W_d)$ be a bisimulation-contracted C-S5$_n$-state, with $M=(W,R,V)$. Then, $|W|$ is bounded in $n$ and $|\atomSet|$.

        \begin{proof}
            Given $k \geq 0$ and a world $w \in W$, we define its \emph{$k$-characteristic formula} $\chi_w^k$ as in \cite{books/el/Goranko2007}:
            \begin{equation*}
                \chi_w^k =
                \begin{cases}
                    L_w,                                                  & \textnormal{if } k=0 \\
                    L_w \wedge
                        \bigwedge\limits_{i \in \agentSet}\left(
                            \textnormal{forth}_{w,i}^k \wedge
                            \textnormal{back}_{w,i}^k
                        \right)                                           & \textnormal{otherwise}
                \end{cases},
            \end{equation*}
            where:
            \begin{align*}
                L_w                        & = \bigwedge_{p \mid w \in V(p)} p \wedge \bigwedge_{p \mid w \not\in V(p)} \neg p \\
                \textnormal{forth}_{w,i}^k & = \bigwedge_{w' \mid w R_i w'} \D{i} \chi_{w'}^{k-1} \\
                \textnormal{back}_{w,i}^k  & = \B{i} \bigvee_{w' \mid w R_i w'} \chi_{w'}^{k-1}
            \end{align*}
            
            We recall the following well-known result from the literature \cite[Theorem 32]{books/el/Goranko2007}:
            
            \begin{appclaim*}[$\star$]
                The following statements are equivalent
                \begin{enumerate}
                    \item $(M, w) \models \chi_v^k$;
                    \item $w \bisim_k v$.
                \end{enumerate}
            \end{appclaim*}
            
            By using Claim ($\star$), Lemma \ref{lem:bounded-bisim-proof} and minimality of models w.r.t. bisimulation, we obtain that for any $w,v \in W$, it holds:
            \begin{equation*}
                (M, w) \models \chi_v^{n+1} \Leftrightarrow w \bisim_{n+1} v \Leftrightarrow w = v.
            \end{equation*}
            
            \noindent Clearly, the size of the set $\{\chi_w^{n+1} \mid w \in W\}$ is bounded in $n$ and $|\atomSet|$, and, hence, the number of worlds of $W$ is also bounded.
        \end{proof}
    \end{lemma}

    \section{\nameref*{sec:general-comm}}
    \subsection{\nameref*{par:b-comm}}
        In this section, we give the full proofs of Theorems \ref{th:dec-b-2} and \ref{th:undec-b-n} of Section \ref{sec:general-comm}.
        
        \subsubsection{Proof of Theorem \ref{th:dec-b-2}}
    To prove Theorem \ref{th:dec-b-2}, we first show some propaedeutical results (Lemma \ref{lem:ck-2-b}, Theorem \ref{th:ck-b}, Corollary \ref{cor:diameter-b} and Lemmata \ref{lem:bounded-bisim-b}, \ref{lem:char-formulae-b}).
    Notice that we follow step by step the proof of Theorem \ref{th:dec} and we give the corresponding results in the logic C$^b$-S5$_2$, for any $b>1$. Since we consider the specific case involving 2 agents, we fix $\agentSet = \{0,1\}$.

    The following is the corresponding version of Lemma \ref{lem:ck-n} of Section \ref{sec:decidability}.

    \begin{applemma}\label{lem:ck-2-b}
        Let $i,j \in \agentSet$ with $i \not= j$, let $ \vec{v} \in \agentSet^* $ and let $\varphi$ be any formula. Then, for any $b{>}1$, in the logic C$^b$-S5$_2$ the formula $ (\B{i} \B{j})^b \varphi \rightarrow \B{v_1} \cdots \B{v_{|\vec{v}|}} \varphi $ is a theorem.

        \begin{proof}
            The proof is by induction on $|\vec{v}|$.
            For the base case ($|\vec{v}|=0$) we have that the formulae $(\B{i}\B{j})^a \varphi \rightarrow \B{j}(\B{i}\B{j})^{a-1} \varphi$ and $\B{j}(\B{i}\B{j})^{a-1} \varphi \rightarrow (\B{i}\B{j})^{a-1} \varphi$ (for each $1 \leq a \leq b$) are instances of \axiom{T}. Together with propositional reasoning, we get that $ (\B{i} \B{j})^b \varphi \rightarrow \varphi $ is a theorem.

            Let now $|\vec{v}| = \lambda$ and suppose by induction that $(\B{i} \B{j})^b \varphi \rightarrow \B{v_1} \cdots \B{v_{\lambda}} \varphi$ is a theorem (for any formula $\varphi$). We now show that, for each $ k \in \agentSet $, the formula $(\B{i} \B{j})^b \varphi \rightarrow \B{v_1} \cdots \B{v_{\lambda}} \B{k} \varphi$ is also a theorem. By inductive hypothesis, substituting $\varphi$ with $\B{k} \varphi$, the following is a theorem:
            \begin{equation*}
                (\B{i} \B{j})^b \B{k} \varphi \rightarrow \B{v_1} \cdots \B{v_{\lambda}} \B{k} \varphi
            \end{equation*}

            \noindent There are now two cases: either $k=j$, or $k=i$. In the former case, we use Equation \ref{eq:box-absoption} as in the proof of Lemma \ref{lem:ck-n-proof} to rewrite the antecedent of the above implication as follows: $ (\B{i} \B{j})^b \B{j} \varphi \equiv (\B{i} \B{j})^{b-1} \B{i} \B{j} \B{j} \varphi \equiv (\B{i} \B{j})^{b-1} \B{i} \B{j} \varphi \equiv (\B{i} \B{j})^b \varphi $.

            In the latter case, we notice that in the logic C$^b$-S5$_2$, for any formula $\varphi$, the following formula is a theorem:
            \begin{equation}\label{eq:comm-2-b}
                (\B{i} \B{j})^b \varphi \leftrightarrow (\B{j} \B{i})^b \varphi
            \end{equation}

            \noindent This immediately follows from axiom \axiom{C$^b$}.

            By using Equation \ref{eq:comm-2-b} we obtain: $ (\B{i} \B{j})^b \B{i} \varphi \equiv (\B{j} \B{i})^b \B{i} \varphi $. By repeating the manipulation of the former case and subsequently reapplying Equation \ref{eq:comm-2-b}, we get: $(\B{j} \B{i})^b \B{i} \varphi \equiv (\B{j} \B{i})^b \varphi \equiv (\B{i} \B{j})^b \varphi$.
            Thus, for each $ k \in \agentSet $, we obtain that the following is a theorem:
            \begin{equation*}
                (\B{i} \B{j})^b \varphi \rightarrow \B{v_1} \cdots \B{v_{\lambda}} \B{k} \varphi.
            \end{equation*}
            This is the required result.
        \end{proof}
    \end{applemma}
    
    The following is the corresponding version of Theorem \ref{th:ck} of Section \ref{sec:decidability}.

    \begin{apptheorem}\label{th:ck-b}
        Let $i,j \in \agentSet$ with $i \not= j$ and let $\varphi$ be any formula. Then, for any $b{>}1$, in the logic C$^b$-S5$_2$ the formula $ (\B{i} \B{j})^b \varphi \leftrightarrow \CK{\agentSet} \varphi $ is a theorem.
        \begin{proof} 
            ($\Leftarrow$) This follows by definition of common knowledge; ($\Rightarrow$) this immediately follows by Lemma \ref{lem:ck-2-b}.
        \end{proof}
    \end{apptheorem}

    The following is the corresponding version of Corollary \ref{cor:diameter} of Section \ref{sec:decidability}.

    \begin{appcorollary}\label{cor:diameter-b}
        Let $i,j \in \agentSet$ with $i \not= j$, let $ \vec{v} \in \agentSet^* $ and let $\varphi$ be any formula. Then, for any $b{>}1$, in an C$^b$-S5$_2$-model we have that if $w R_{v_1} \circ \ldots \circ R_{v_{|\vec{v}|}} w'$, then $w (R_{i} \circ R_{j})^b w'$.
    \end{appcorollary}

    The statement above directly follows from the contrapositive of the implication in Lemma \ref{lem:ck-2-b}, under the assumption of minimality of models (w.r.t. bisimulation).
    Intuitively, this states that if a world of a C$^b$-S5$_2$-model is reachable in an arbitrary number of steps, then it is also reachable in exactly $2b$ steps.

    The following is the corresponding version of Lemma \ref{lem:bounded-bisim} of Section \ref{sec:decidability}.

    \begin{applemma}\label{lem:bounded-bisim-b}
        Let $(M, W_d)$ be an C$^b$-S5$_2$-state, with $M=(W,R,V)$. For any $w,v \in W$, we have that $ w \bisim_{2b+1} v \iff w \bisim v$.

        \begin{proof}
            The proof is identical to that of Lemma \ref{lem:bounded-bisim}, by using Corollary \ref{cor:diameter-b} instead of Corollary \ref{cor:diameter}.
        \end{proof}
    \end{applemma}

    The following is the corresponding version of Lemma \ref{lem:char-formulae} of Section \ref{sec:decidability}.

    \begin{applemma}\label{lem:char-formulae-b}
        Let $(M,W_d)$ be an C$^b$-S5$_2$-state, with $M=(W,R,V)$. Then, $|W|$ is bounded in $2b$ and $|\atomSet|$.

        \begin{proof}
            The proof is identical to that of Lemma \ref{lem:char-formulae}.
        \end{proof}
    \end{applemma}

    \settheoremcountertoref{th:dec-b-2}
    \begin{thm}\label{th:dec-b-2-proof}
        For any $b{>}1$, \planex{$\mathcal{T}_{\textnormal{C$^b$-S5}}$}{$2$} is \emph{decidable}.

        \begin{proof}
            Let $T \in \mathcal{T}_{\textnormal{C}^b-\textnormal{S5}_2}$ be an epistemic planning task. By Lemma \ref{lem:char-formulae-b}, it follows that we can perform a breadth-first search on the search space that would only visit a finite number of epistemic states (up to bisimulation contraction) to find a solution for $T$. Thus, we obtain the claim.
        \end{proof}
    \end{thm}
    
        \newcommand{\inc}[1]{\textnormal{inc}(#1)}
\newcommand{\jump}[1]{\textnormal{jump}(#1)}
\newcommand{\jzdec}[2]{\textnormal{jzdec}(#1,#2)}
\newcommand{\halt}{\textnormal{halt}}

\subsubsection{Proof of Theorem \ref{th:undec-b-n}}
    To prove Theorem \ref{th:undec-b-n}, we first show some propaedeutical results (Lemmata \ref{lem:index-meta-chain}, \ref{lem:index-operations}, \ref{lem:comp-function}, \ref{lem:halting}).

    In what follows, we consider the case with $\agentSet = \{0,1,2\}$ and $b=2$, \ie we focus on the logic C$^2$-S5$_3$. Since C$^2$-S5$_3$-models are also C$^b$-S5$_n$-models for any $n > 3$ and $b > 2$, our results hold for any combination of the values of $n \geq 3$ and $b \geq 2$. Moreover, we fix $\atomSet = \{p_1,p_2,p_3,r\}$.
    
    We adapt the proof in \cite[Section~6]{conf/ijcai/Aucher2013}, of the undecidability of epistemic planning in the logic $S5_n$ ($n > 1$). It is an elegant reduction from the problem of reachability in Minsky two-counter machines to the problem of epistemic planning. We first, recall the definition.

    \begin{appdefinition}[Two-counter machine]
        A \emph{two-counter machine} $M$ is a finite sequence of instructions $(I_0, \dots, I_T$), where each instruction $I_t$, with $t<T$, is from the set:
        \begin{equation*}
            \{\inc{i}, \jump{j}, \jzdec{i}{j} \mid i = 0,1, j \leq T\},
        \end{equation*}
        and $I_T=\halt$. A \emph{configuration} of $M$ is a triple $(k,l,m) \in \mathbb{N}^3$, where $k$ is the index of the current instruction, and $l$ and $m$ are the current contents of counters 0 and 1, respectively. The \emph{computation function} $f_M : \mathbb{N} \rightarrow \mathbb{N}^3$ of $M$ maps time steps into configurations, ad is given by $f_M(0) = (0,0,0)$ and if $f_M(n) = (k,l,m)$, then:
        \begin{equation*}
            f_M(n{+}1){=}
            \begin{cases}
                (k{+}1,l{+}1,m    ) & \textnormal{if } I_k {=} \inc{0}                                 \\
                (k{+}1,l    ,m{+}1) & \textnormal{if } I_k {=} \inc{1}                                 \\
                (j    ,l    ,m    ) & \textnormal{if } I_k {=} \jump{j}                                \\
                (j    ,l    ,m    ) & \textnormal{if } I_k {=} \jzdec{0}{j} \textnormal{ and } l {=} 0 \\
                (j    ,l    ,m    ) & \textnormal{if } I_k {=} \jzdec{1}{j} \textnormal{ and } m {=} 0 \\
                (k{+}1,l{-}1,m    ) & \textnormal{if } I_k {=} \jzdec{0}{j} \textnormal{ and } l {>} 0 \\
                (k{+}1,l    ,m{-}1) & \textnormal{if } I_k {=} \jzdec{1}{j} \textnormal{ and } m {>} 0 \\
                (k    ,l    ,m    ) & \textnormal{if } I_k {=} \halt
            \end{cases}
        \end{equation*}
        We say that $M$ \emph{halts} if $f_M(n) {=} (T,l,m)$ for some $n,l,m {\in} \mathbb{N}$.
    \end{appdefinition}

    \begin{apptheorem}[\cite{book/ph/Minsky1967}]\label{th:minsky}
        The halting problem for two-counter machines is undecidable.
    \end{apptheorem}

    We follow the approach of \cite{conf/ijcai/Aucher2013} step by step by encoding the halting problem of a Minsky machine $M$ as an epistemic planning task. The procedure follows three steps:
    \begin{enumerate}
        \item We define an encoding for integers and configurations;
        \item We build a finite set of actions for encoding the computation function $f_M$; and
        \item We combine the previous steps and we encode the halting problem as an epistemic planning task.
    \end{enumerate}

    In all figures, reflexive, transitive (and symmetric) edges are implicit. In the models, the worlds are labelled with the name of the world and the propositions true in it. In the event models, the events are labelled with the name of the event and the precondition; there are no postconditions.

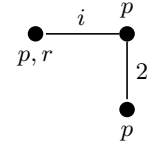
\begin{figure}[ht]
        \centering
        \begin{tikzpicture}[-,>=stealth',shorten >=1pt,auto,node distance=2cm, semithick]

  \node (A1) [world,label=above:$p$] {};
  \node (A2) [world,below=0.8cm of A1,label=below:{$p$}] {};

  \node (B1) [world,left=1cm of A1,label=below:{$p,r$}] {};

  \path (A1)
  edge              node [above] {$i$} (B1)
  edge              node [right] {$2$} (A2);

\end{tikzpicture}
        \caption{\label{fig:metaworld}\METAWORLD$_i(p)$, with $0 \leq i \leq 1$.}
    \end{figure}

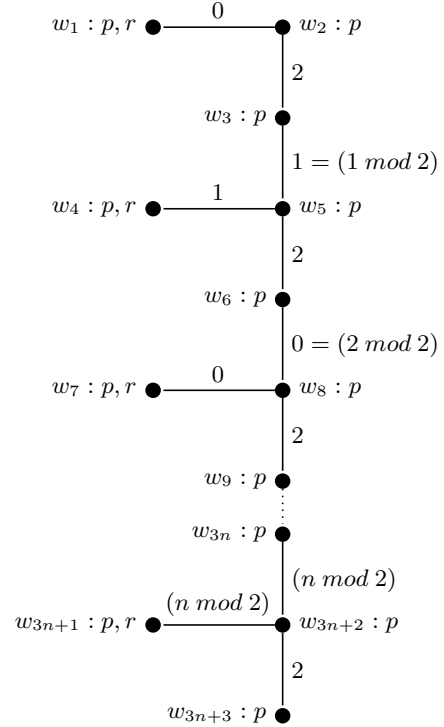
\begin{figure}[ht]
        \centering
        \begin{tikzpicture}[-,>=stealth',shorten >=1pt,auto,node distance=2cm, semithick]

  \node (A1) [world,                 label=right:{$w_2: p$}] {} ;
  \node (A2) [world,below=1cm of A1, label=left: {$w_3: p$}] {} ;
  \node (A3) [world,below=1cm of A2, label=right:{$w_5: p$}] {} ;
  \node (A4) [world,below=1cm of A3, label=left: {$w_6: p$}] {} ;
  \node (A5) [world,below=1cm of A4, label=right:{$w_8: p$}] {} ;
  \node (A6) [world,below=1cm of A5, label=left: {$w_9: p$}] {} ;
  \node (A7) [world,below=.5cm of A6, label=left: {$w_{3n}: p$}] {} ;
  \node (A8) [world,below=1cm of A7, label=right:{$w_{3n+2}: p$}] {} ;
  \node (A9) [world,below=1cm of A8, label=left: {$w_{3n+3}: p$}] {} ;

  \node (B1) [world,left=1.5cm of A1, label=left:{$w_1: p, r$}] {} ;
  \node (B3) [world,left=1.5cm of A3, label=left:{$w_4: p, r$}] {} ;
  \node (B5) [world,left=1.5cm of A5, label=left:{$w_7: p, r$}] {} ;
  \node (B8) [world,left=1.5cm of A8, label=left:{$w_{3n+1}: p, r$}] {} ;


  \path (A1)
  edge              node [above] {$0$} (B1)
  edge              node [right] {$2$} (A2);

  \path (A2)
  edge              node [right] {$1 = (1~mod~2)$} (A3);

  \path (A3)
  edge              node [above] {$1$} (B3)
  edge              node [right] {$2$} (A4);
  
  \path (A4)
  edge              node [right] {$0 = (2~mod~2)$} (A5);
  
  \path (A5)
  edge              node [above] {$0$} (B5)
  edge              node [right] {$2$} (A6);
  
   \path (A6)
    edge[dotted]      node [right] {} (A7);

   \path (A7)
    edge      node [right] {$(n~mod~2)$} (A8); 
  
  \path (A8)
  edge              node [above] {$(n~mod~2)$} (B8)
  edge              node [right] {$2$} (A9);

    
    
    

\end{tikzpicture}
        \caption{\label{fig:metachain}\METACHAIN$(p,n)$: Meta-chains are chains of alternating meta-worlds linked alternatingly by a relation~$1$ and a relation~$0$. \METACHAIN$(p,n)$ is a chain of $n+1$ \METAWORLD$(p)$.}
    \end{figure}

    \boldparagraph{Integers and configurations.}
    For each $p \in \atomSet$ and each $n \in \mathbb{N}$, we define an epistemic model \METACHAIN$(p,n)$, represented in Figure~\ref{fig:metachain}, which contains $n+1$ meta-worlds (models themselves, Figure~\ref{fig:metaworld}). Thus, the integer $0$ is represented by the meta-chain made of only the meta-world model of Figure~\ref{fig:metachain-p-0}.
    Finally, for each configuration $(k,l,m) \in \mathbb{N}^3$ of two-counter machines, we define the epistemic model \METASTATE$_{(k,l,m)}$ as in Figure~\ref{fig:meta-state}.

    In \cite{conf/ijcai/Aucher2013}, the meta-worlds that compose a meta-chain are always linked together with the same accessibility relation. As a consequence, a meta-chain can have an arbitrary long series of alternating distinct worlds $u_1 \overset{i}{\rightarrow} u_2 \overset{j}{\rightarrow} u_3 \overset{i}{\rightarrow} u_4 \overset{j}{\rightarrow} u_5 \cdots$ with $i$ and $j$ distinct and all $u_k$ distinct.
    This is not possible in the logic C$^2$-S5$_n$, due to axiom \axiom{C$^2$}.
    Thus, we need to `bypass' axiom \axiom{C$^2$} in the meta-chains. To do so, we devised meta-chains so that meta-worlds are alternatingly linked together with two different relations. This difference also forces us to use three agents instead of two (like in \cite{conf/ijcai/Aucher2013}) and, in fact, the plan existence problem in the two agents case is decidable (see Theorem \ref{th:dec-b-2}).

    Notice how in a meta-state, for $i \not = j$, the longest series of alternating distinct worlds $u_1 \overset{i}{\rightarrow} u_2 \overset{j}{\rightarrow} u_3 \overset{i}{\rightarrow} u_4 \overset{j}{\rightarrow} u_5 \cdots$ with all $u_k$ distinct, is bounded, and is 4. Thus, the models are thus vacuously models of C$^2$-S5$_3$.

    \begin{figure}[ht]
        \centering
        \begin{tikzpicture}[-,>=stealth',shorten >=1pt,auto,node distance=2cm, semithick]

  \node (A1) [world,                 label=right:{$w_2: \mu_0(p)$}] {} ;
  \node (A2) [world,below=1cm of A1, label=below:{$w_3: \lambda_0(p)$}] {} ;
  \node (B1) [world,left=1cm  of A1, label=left: {$w_1: \tau_0(p)$}] {} ;

  
  \path (A1)
  edge              node [above] {$0$} (B1)
  edge              node [right] {$2$} (A2);

\end{tikzpicture}
        \caption{\label{fig:metachain-p-0}\METACHAIN$(p,0)$}
    \end{figure}

    \begin{figure}[ht]
        \centering
        \begin{tikzpicture}[sibling distance=3cm,semithick]

  \node (root) [pointedworld,label={$w_0$}] {}
    child { node [inner sep=0,anchor=south east] {} 
    {node (l) [itria] {\scalebox{0.8}{\rotatebox{90}{\METACHAIN$(p_1, k)$}}}} edge from parent node[above,draw=none] {$0$}
    }
    child { node [inner sep=0,anchor=south] {}
    {node (m) [itria] {\scalebox{0.8}{\rotatebox{90}{\METACHAIN$(p_2, l)$\hspace{0.2cm}}}} } edge from parent node[left,draw=none] {$0$} 
    }
    child { node [inner sep=0,anchor=south west] {}
    {node (r) [itria] {\scalebox{0.8}{\rotatebox{90}{\METACHAIN$(p_3, m)$}}} } edge from parent node[above,draw=none] {$0$} 
    }
  ;
\end{tikzpicture}
        \caption{\label{fig:meta-state} \METASTATE$_{(k,l,m)}$}
    \end{figure}


    \boldparagraph{Computation function.}
    First, we define path formulae.

    \begin{appdefinition}[Path formulae]
        For all $p \in \atomSet$ and $n \in \mathbb{N}$, we inductively define the formulae $\lambda_0(p), \mu_0(p), \tau_0(p)$ as follows:
        \begin{itemize}
            \item $\lambda_0(p) = p \land \B{0} \lnot r \land \B{1} \lnot r$
            \item $\mu_0(p) = p \land \D{2} \lambda_0(p) \land \neg \lambda_0(p)$
            \item $\tau_0(p) = p \land r \land (\D{0} \mu_0(p) \lor \D{1} \mu_0(p))$

            \item $\lambda_{n+1}(p) = p \land \lnot r \land \lnot \mu_n(p) \land (\Diamond_0 \mu_n(p) \lor \Diamond_1 \mu_n(p))$
            \item $\mu_{n+1}(p) = p \land \D{2} \lambda_{n+1}(p) \land \neg \lambda_{n+1}(p)$
            \item $\tau_{n+1}(p) = p \land r \land (\D{0} \mu_{n+1}(p) \lor \D{1} \mu_{n+1}(p))$
            %
        \end{itemize}
    \end{appdefinition}

    \begin{applemma}\label{lem:index-meta-chain}
        For all $p {\in} \atomSet$, $n {\in} \mathbb{N}$, $0 \leq i \leq n$, $1 \leq j \leq 3n{+}3$:
        \begin{equation*}
            \begin{array}{lll}
                (\METACHAIN(p,n), w_j) \models \lambda_i(p) & {\Leftrightarrow} & j {=} 3(n{-}i) {+} 3 \\
                (\METACHAIN(p,n), w_j) \models \mu_i(p)     & {\Leftrightarrow} & j {=} 3(n{-}i) {+} 2 \\
                (\METACHAIN(p,n), w_j) \models \tau_i(p)    & {\Leftrightarrow} & j {=} 3(n{-}i) {+} 1
            \end{array}
        \end{equation*}
    \end{applemma}
    That is, the path formulas allow one to uniquely identify worlds in a meta-chain.
    In the $(i + 1)th$ to last meta-world in META-CHAIN$(p,n)$, 
    $\lambda_i(p)$ holds in the bottom world,
    $\mu_i(p)$ in the top-right world, 
    $\tau_i(p)$ in the top-left world.
    Figure~\ref{fig:metachain-p-0} illustrates the base cases.

    The instructions of a two-counter machine can be decomposed in simple operations on integers: \emph{increment}, \emph{decrement} and \emph{replacement}. We encode each operation with an event model, represented on Figures \ref{fig:metainc}, \ref{fig:metadec} and \ref{fig:metarepl}, respectively. Due to the structure of meta-chains, that comprise alternating accessibility relations, we need to define two different event models for increment. Namely, \METAINC$_0(p)$ is used to increment odd numbers and \METAINC$_1(p)$ handles even numbers. Thus, given an integer $n$, to increment \METACHAIN$(p,n)$, we use \METAINC$_i(p)$, where $i = 1-(n~mod~2)$.

    The following Lemma makes sure that the operations on integers are correctly captured by the product update of meta-chains with the event models for increment, decrement and replacement.

    \begin{applemma}\label{lem:index-operations}
        For all $m,n \in \mathbb{N}$ and for all $p \in \atomSet$, we have:
        \begin{enumerate}
            \item\label{item:inc}  $\METACHAIN(p, n) \otimes \METAINC_i(p) =\\ \METACHAIN(p, n+1)$, where $i = 1-(n~mod~2)$;
            \item\label{item:dec}  If $n > 0$, $\METACHAIN(p, n) \otimes \METADEC(p) =\\ \METACHAIN(p, n - 1)$;
            \item\label{item:repl} $\METACHAIN(p, n) \otimes \METAREPL(p, n, m) =\\ \METACHAIN(p, m)$.
        \end{enumerate}

        \begin{proof}
            First, we consider item \ref{item:inc}, \ie event model of Figure \ref{fig:metainc}. Let $n$ be even, \ie $i=1$ (the case with $i=0$ is identical). The top event of Figure \ref{fig:metainc} is paired with all worlds in $\METACHAIN(p, n)$, except for the one where $\lambda_0(p)$ holds. From Lemma \ref{lem:index-meta-chain}, this world in unique and it is the bottom world of $\METACHAIN(p, n)$. Thus, after the product of $\METACHAIN(p, n)$ with the top event, we obtain a copy of the chain, except for its bottom world. Such world is paired with the second-to-top event of Figure \ref{fig:metainc}. At this point, we obtain an exact copy of $\METACHAIN(p, n)$. Finally, the last three events of Figure \ref{fig:metainc} create an additional meta-world. Since $n$ is even, the last relation linking meta-worlds in the chain is that of agent $0$. Then, the event model $\METAINC_1(p)$ links the additional meta-world to the bottom of chain with the accessibility relation of agent $1$. Thus, we obtain $\METACHAIN(p, n+1)$.

            We now focus on item \ref{item:dec}, \ie event model of Figure \ref{fig:metadec}. By Lemma \ref{lem:index-meta-chain}, its only event is paired with all worlds of $\METACHAIN(p, n)$, except for those in the bottom meta-world. Since $n>0$, we obtain $\METACHAIN(p, n-1)$.

            Finally, we consider item \ref{item:repl}, \ie event model of Figure \ref{fig:metarepl}. By Lemma \ref{lem:index-meta-chain}, the $2 \cdot (m+1)$ event models on the right hand side of Figure \ref{fig:metarepl} all pair with the top-right world of the top meta-world of $\METACHAIN(p, n)$ and the $m+1$ events on the left hand side pair with the top left world of the same meta-world. Thus, we create $m+1$ copies of the top meta-world of $\METACHAIN(p, n)$. Then, these copies are alternatingly linked together with the accessibility relations of agents $0$ and $1$. Thus, we obtain $\METACHAIN(p, m)$.
        \end{proof}
    \end{applemma}

    \begin{figure}
        \centering
        \begin{tikzpicture}[-,>=stealth',shorten >=1pt,auto,node distance=2cm, semithick]

  \node (A0) [event,                   label=right:{$p \land \lnot \lambda_0(p)$}] {} ;
  \node (A1) [event,below=1cm of A0, label=right:{$p \land \mu_0(p)$}] {} ;

  \node (A2) [event,below=1cm of A1, label=right:{$p \land \mu_0(p)$}] {} ;
  \node (A3) [event,below=1cm of A2, label=right:{$p \land \mu_0(p)$}] {} ;
  
  \node (B1) [event,left=1cm  of A2, label=left:{$p \land \tau_0(p)$}] {} ;


  \path (A0)
  edge              node [right] {$2$} (A1) ;

  \path (A1)
  edge              node [right] {$i$} (A2) ;

  \path (A2)
  edge              node [above] {$i$} (B1)
  edge              node [right] {$2$} (A3);



  





\end{tikzpicture}
        \caption{\label{fig:metainc}\METAINC$_i(p)$, with $0 \leq i \leq 1$.}
    \end{figure}
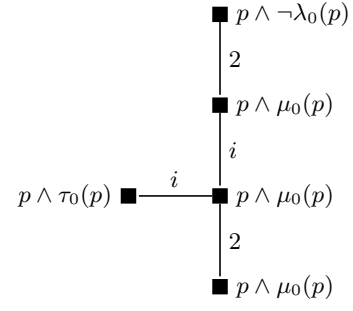 

    \begin{figure}
        \centering
        \begin{tikzpicture}[-,>=stealth',shorten >=1pt,auto,node distance=2cm, semithick]

  \node (A1) [event,label=below:{$p \land \lnot\lambda_0(p) \land \lnot\mu_0(p) \land \lnot\tau_0(p)$}] {} ;
\end{tikzpicture}
        \caption{\label{fig:metadec}\METADEC$(p)$}
    \end{figure}

    \begin{figure}
        \centering
        \begin{tikzpicture}[-,>=stealth',shorten >=1pt,auto,node distance=2cm, semithick]

  \node (A1) [event,                 label=right:{$p \land \mu_n(p)$}] {} ;
  \node (A2) [event,below=1cm of A1, label=left:{$p \land \mu_n(p)$}] {} ;
  \node (A3) [event,below=1cm of A2, label=right:{$p \land \mu_n(p)$}] {} ;
  \node (A4) [event,below=1cm of A3, label=left:{$p \land \mu_n(p)$}] {} ;
  \node (A5) [event,below=1cm of A4, label=right:{$p \land \mu_n(p)$}] {} ;
  \node (A6) [event,below=1cm of A5, label=left:{$p \land \mu_n(p)$}] {} ;
  \node (A7) [event,below=.5cm of A6, label=left:{$p \land \mu_n(p)$}] {} ;
  \node (A8) [event,below=1cm of A7, label=right:{$p \land \mu_n(p)$}] {} ;
  \node (A9) [event,below=1cm of A8, label=left:{$p \land \mu_n(p)$}] {} ;

  \node (B1) [event,left=2cm  of A1, label=left:{$p \land \tau_n(p)$}] {} ;
  \node (B3) [event,left=2cm  of A3, label=left:{$p \land \tau_n(p)$}] {} ;
  \node (B5) [event,left=2cm  of A5, label=left:{$p \land \tau_n(p)$}] {} ;
  \node (B8) [event,left=2cm  of A8, label=left:{$p \land \tau_n(p)$}] {} ;


  \path (A1)
  edge              node [above] {$0$} (B1)
  edge              node [right] {$2$} (A2);

  \path (A2)
  edge              node [right] {$1 = (1~mod~2)$} (A3);

  \path (A3)
  edge              node [above] {$1 = (1~mod~2)$} (B3)
  edge              node [right] {$2$} (A4);
  
  \path (A4)
  edge              node [right] {$0 = (2~mod~2)$} (A5);
  
  \path (A5)
  edge              node [above] {$0 = (2~mod~2)$} (B5)
  edge              node [right] {$2$} (A6);
  
   \path (A6)
    edge[dotted]      node [right] {} (A7);

   \path (A7)
    edge      node [right] {$(m~mod~2)$} (A8); 
  
  \path (A8)
  edge              node [above] {$(m~mod~2)$} (B8)
  edge              node [right] {$2$} (A9);

    
    
    

\end{tikzpicture}
        \caption{\label{fig:metarepl}\METAREPL$(p,n,m)$}
    \end{figure}

    For all $k \in \mathbb{N}$, we define $\phi_k = \D{0}\mu_k(p_1)$. By Lemma \ref{lem:index-meta-chain} and the definition of \METASTATE$_{(k,l,m)}$, we immediately obtain that for all $k,l,m,k' \in \mathbb{N}$ the following holds:
    \begin{equation}\label{eq:meta-s}
        \METASTATE_{(k,l,m)} \models \phi_{k'} \textnormal{ iff } k' = k.
    \end{equation}

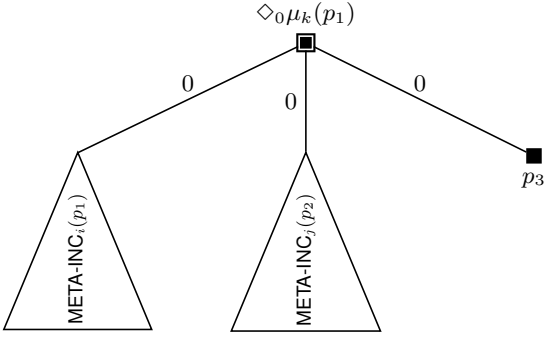
\begin{figure}[ht]
        \begin{tikzpicture}[sibling distance=3cm,semithick]

  \node (root) [pointedevent,label={$\Diamond_0\mu_k(p_1)$}] {}
    child { node [inner sep=0,anchor=south east] {} 
    {node (l) [itria] {\scalebox{0.8}{\rotatebox{90}{\METAINC$_i(p_1)$}}}} edge from parent node[above,draw=none] {$0$}
    }
    child { node [inner sep=0,anchor=south] {}
    {node (m) [itria] {\scalebox{0.8}{\rotatebox{90}{\METAINC$_j(p_2)$}}} } edge from parent node[left,draw=none] {$0$} 
    }
    child { node [event,label=below:{$p_3$}] {} edge from parent node[above,draw=none] {$0$} 
    }
  ;
\end{tikzpicture}
        \caption{\label{fig:action-inc} The action $a_M(k,l,m)$ when $I_k = \inc{0}$, $i = 1-(k~mod~2)$ and $j = 1-(l~mod~2)$. The case $I_k = \inc{1}$ is obtained by replacing $p_2$ and $p_3$ everywhere and by having $j = 1-(m~mod~2)$.}
    \end{figure}

\begin{figure}[ht]
        \centering
        \begin{tikzpicture}[sibling distance=3cm,semithick]

  \node (root) [pointedevent,label={$\D{0} \mu_k(p_1) \land \D{0} \mu_0(p_2)$}] {}
    child { node [inner sep=0,anchor=south east] {} 
    {node (l) [itria] {\scalebox{0.8}{\rotatebox{90}{\METAREPL$(p_1, k, j)$}}}} edge from parent node[above,draw=none] {$0$}
    }
    child { node [event,label=below:{$p_2$}] {}
     edge from parent node[left,draw=none] {$0$} 
    }
    child { node [event,label=below:{$p_3$}] {}
     edge from parent node[above,draw=none] {$0$} 
    }
  ;
\end{tikzpicture}
        \caption{\label{fig:action-jzdec-z} The action $a_M(k,l,m)$ when $I_k = \jzdec{0}{j}, l=0$. The case $I_k = \jzdec{1}{j}, m=0$ is obtained by replacing $p_2$ and $p_3$ in the precondition of the designated event.}
    \end{figure}
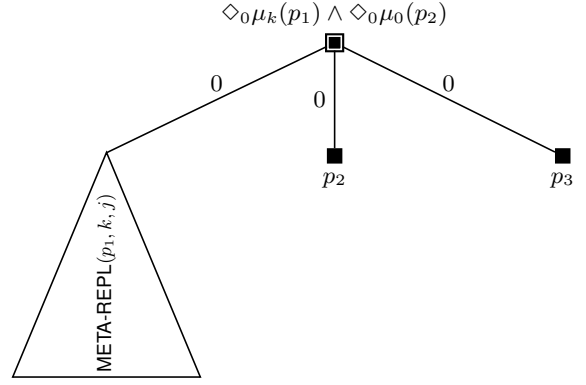

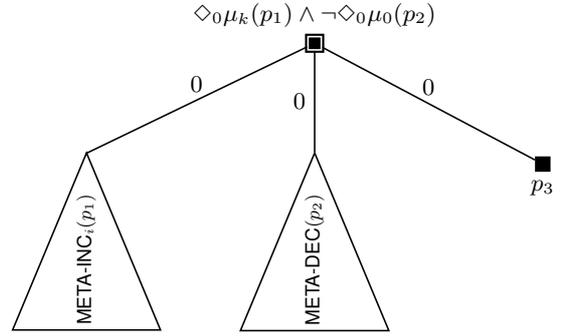
\begin{figure}[ht]
        \centering
        \begin{tikzpicture}[sibling distance=3cm,semithick]

  \node (root) [pointedevent,label={$\D{0} \mu_k(p_1) \land \lnot \D{0} \mu_0(p_2)$}] {}
    child { node [inner sep=0,anchor=south east] {} 
    {node (l) [itria] {\scalebox{0.8}{\rotatebox{90}{\METAINC$_i(p_1)$}}}} edge from parent node[above,draw=none] {$0$}
    }
    child { node [inner sep=0,anchor=south] {}
    {node (l) [itria] {\scalebox{0.8}{\rotatebox{90}{\METADEC$(p_2)$}}}} 
     edge from parent node[left,draw=none] {$0$} 
    }
    child { node [event,anchor=north,label=below:{$p_3$}] {}
     edge from parent node[above,draw=none] {$0$} 
    }
  ;
\end{tikzpicture}
        \caption{\label{fig:action-jzdec} The action $a_M(k,l,m)$ when $I_k = \jzdec{0}{j}$, $l>0$ and $i = 1-(k~mod~2)$. Case $I_k = \jzdec{1}{j}$ and $m>0$ is obtained by replacing $p_2$ and $p_3$ everywhere.}
    \end{figure}

    Let now $M = (I_0, \dots, I_T)$ be a two-counter machine. For all $k<T$ and $l,m \in \mathbb{N}$, we define an epistemic action $a_M(k,l,m)$ as in Figures \ref{fig:action-inc}-\ref{fig:action-jzdec}.

    \begin{figure}[ht]   
        \centering
        \begin{tikzpicture}[sibling distance=3cm,semithick]

  \node (root) [pointedevent,label={$\D{0} \mu_k(p_1)$}] {}
    child { node [inner sep=0,anchor=south east] {} 
    {node (l) [itria] {\scalebox{0.8}{\rotatebox{90}{\METAREPL$(p_1, k, j)$}}}} edge from parent node[above,draw=none] {$0$}
    }
    child { node [event,label=below:{$p_2$}] {}
     edge from parent node[left,draw=none] {$0$} 
    }
    child { node [event,label=below:{$p_3$}] {}
     edge from parent node[above,draw=none] {$0$} 
    }
  ;
\end{tikzpicture}
        \caption{\label{fig:action-jump} The action $a_M(k,l,m)$ when $I_k = \jump{j}$.}
    \end{figure}
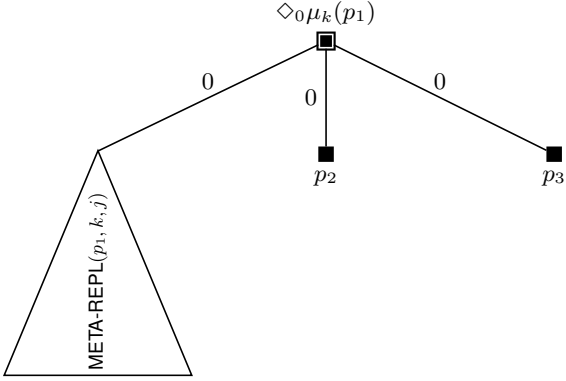

    We now define a notion of \emph{equivalence} between configurations. Two configurations $(k,l,m), (k',l',m') \in \mathbb{N}^3$ are equivalent, denoted by $(k,l,m) \approx (k',l',m')$ if the following holds:
    \begin{equation*}
        k=k' \textnormal{ and }
        \begin{cases}
            l=0 \leftrightarrow l'=0 & \textnormal{if } I_k = \jzdec{0}{j} \\
            m=0 \leftrightarrow m'=0 & \textnormal{if } I_k = \jzdec{1}{j}
        \end{cases}.
    \end{equation*}
    Notice that if $(k,l,m) \approx (k',l',m')$, then $a_M(k,l,m) = a_M(k',l',m')$. Thus, the following set is \emph{finite}:
    \begin{equation*}
        \mathcal{F}_M := \{a_M(k,l,m) \mid 0 \leq k < T \textnormal{ and } l,m \in \mathbb{N}\}.
    \end{equation*}

    The following Lemma shows that $\mathcal{F}_M$ correctly encodes the computation function of the two-counter machine $M$.
    \begin{applemma}\label{lem:comp-function}
        Let $M = (I_0, \dots, I_T)$ be a two-counter machine, $l,m,n \in \mathbb{N}$ and $k<T$. Then, the following holds:
        \begin{enumerate}
            \item $a_M(k,l,m)$ is applicable in $\METASTATE_{f_M(n)}$ iff $(k,l,m) \approx f_M(n)$;
            \item $\METASTATE_{f_M(n)} \otimes a_M(f_M(n)) = \METASTATE_{f_M(n+1)}$.
        \end{enumerate}

        \begin{proof}
            Let $f_M(n) = (k',l',m')$. The first item by case of $I_k$.
            \begin{itemize}
                \item $I_k = \inc{0}$, $\inc{1}$, or $\jump{j}$: $a_M(k,l,m)$ is an action of the form $(\E, \{e\})$ with $pre(e) = \D{0}\mu_k(p_1) = \phi_k$. Thus, by equation \ref{eq:meta-s}, we have: $a_M(k,l,m)$ is applicable in $\METASTATE_{f_M(n)} \Leftrightarrow \METASTATE_{(k',l',m')} \models \phi_k \Leftrightarrow k=k' \Leftrightarrow (k,l,m) \approx (k',l',m')$.
                \item $I_k = \jzdec{0}{j}$ and $l = 0$: $a_M(k,l,m)$ is an action of the form $(\E, \{e\})$ with $pre(e) = \D{0} \mu_k(p_1) \land \D{0} \mu_0(p_2) = \phi_k \land \D{0} \mu_0(p_2)$. Thus, by equation \ref{eq:meta-s}, we have: $a_M(k,l,m)$ is applicable in $\METASTATE_{f_M(n)} \Leftrightarrow \METASTATE_{(k',l',m')} \models \phi_k \land \D{0} \mu_0(p_2)$ $\Leftrightarrow k=k' \land l'=0 \Leftrightarrow (k,l,m) \approx (k',l',m')$.
                \item $I_k = \jzdec{1}{j}$ and $m = 0$: analogous to the previous case.
                \item $I_k = \jzdec{0}{j}$ and $l > 0$: $a_M(k,l,m)$ is an action of the form $(\E, \{e\})$ with $pre(e) = \D{0} \mu_k(p_1) \land \lnot\D{0} \mu_0(p_2) = \phi_k \land \lnot\D{0} \mu_0(p_2)$. Thus, by equation \ref{eq:meta-s}, we have: $a_M(k,l,m)$ is applicable in $\METASTATE_{f_M(n)} \Leftrightarrow \METASTATE_{(k',l',m')} \models \phi_k \land \lnot\D{0} \mu_0(p_2)$ $\Leftrightarrow k=k' \land l'\not=0 \Leftrightarrow (k,l,m) \approx (k',l',m')$.
                \item $I_k = \jzdec{1}{j}$ and $m > 0$: analogous to the previous case.
            \end{itemize}

            The second item is by case of $I_{k'}$:
            \begin{itemize}
                \item $I_{k'} = \inc{0}$: $a_M(k',l',m')$ is the action of Figure \ref{fig:action-inc}. Thus, by Lemma \ref{lem:index-operations}, we have that: $\METASTATE_{f_M(n)} \otimes a_M(f_M(n)) =$ $\METASTATE_{(k',l',m')} \otimes a_M(k',l',m') = \METASTATE_{(k'+1,l'+1,m')} = \METASTATE_{f_M(n+1)}$.
                \item $I_{k'} = \inc{1}$: analogous to the previous case.
                \item $I_{k'} = \jump{j}$: $a_M(k',l',m')$ is the action of Figure \ref{fig:action-jump}. Thus, by Lemma \ref{lem:index-operations}, we have that: $\METASTATE_{f_M(n)} \otimes a_M(f_M(n)) = \METASTATE_{(k',l',m')} \otimes a_M(k',l',m') = \METASTATE_{(j,l',m')} = \METASTATE_{f_M(n+1)}$.
                \item $I_{k'} = \jzdec{0}{j}$ and $l = 0$: $a_M(k',l',m')$ is the action of Figure \ref{fig:action-jzdec-z}. Thus, by Lemma \ref{lem:index-operations}, we have that: $\METASTATE_{f_M(n)} \otimes a_M(f_M(n)) = \METASTATE_{(k',l',m')} \otimes a_M(k',l',m') = \METASTATE_{(j,l',m')} = \METASTATE_{f_M(n+1)}$.
                \item $I_{k'} = \jzdec{1}{j}$ and $m = 0$: analogous to the previous case.
                \item $I_{k'} = \jzdec{0}{j}$ and $l > 0$: $a_M(k',l',m')$ is the action of Figure \ref{fig:action-jzdec}. Thus, by Lemma \ref{lem:index-operations}, we have that: $\METASTATE_{f_M(n)} \otimes a_M(f_M(n)) =$ $\METASTATE_{(k',l',m')} \otimes a_M(k',l',m') = \METASTATE_{(k'+1,l'-1,m')} = \METASTATE_{f_M(n+1)}$.
                \item $I_{k'} = \jzdec{1}{j}$ and $m > 0$: analogous to the previous case.
            \end{itemize}
        \end{proof}
    \end{applemma}


    \boldparagraph{Halting problem.}
    From Lemma \ref{lem:comp-function}, we obtain the following result:

    \begin{applemma}\label{lem:halting}
        Let $M {=} (I_0, \ldots, I_T)$ be a two-counter machine. We define the epistemic planning task $T_M = (\METASTATE_{(0,0,0)}, \mathcal{F}_M,$ $\phi_T)$. Then, $T_M$ has a solution iff $M$ halts.

    \end{applemma}

    Thus, from Lemma \ref{lem:halting} and Theorem \ref{th:minsky} and from the fact that C$^2$-S5$_3$-models are also C$^b$-S5$_n$-models for any $n > 3$ and $b > 2$, we obtain:
    
    \settheoremcountertoref{th:undec-b-n}
    \begin{thm}\label{th:undec-b-n-proof}
        For any $n{>}2$ and $b{>}1$, \planex{$\mathcal{T}_{\textnormal{C$^b$-S5}}$}{$n$} is \emph{undecidable}.
    \end{thm}
    
    We summarize the results of this section in Table~\ref{tab:summary}.

    \begin{table}[h]
        \centering
        \begin{tabular}{l|cccc}
            \toprule
            $\downarrow n$ / $b \rightarrow$ & 1 & 2 & 3 & \ldots \\
            \midrule
            1 & D & D & D & D\\
            2 & D & D & D & D \\
            3 & D & UD & UD & UD \\
            4 & D & UD & UD & UD \\
            \ldots & D & UD & UD & UD \\
            \bottomrule
        \end{tabular}
    \caption{\label{tab:summary} Summary of complexity results for the \planex{$\mathcal{T}_{\textnormal{C$^b$-S5}}$}{$n$}. D: decidable; UD: undecidable.}
    \end{table}

    \subsection{\nameref*{par:weak-comm}}
        In this section, we give the full proofs of Theorem \ref{th:dec-l} of Section \ref{sec:general-comm}.

        \subsubsection{Proof of Theorem \ref{th:dec-l}}
    To prove Theorem \ref{th:dec-l}, we first show some propaedeutical results (Lemmata \ref{lem:slide-box-l}, \ref{lem:ck-n-l}, Theorem \ref{th:ck-l} and Corollary \ref{cor:diameter-l}).

    The following is the corresponding version of Lemma \ref{lem:slide-box} of Section \ref{sec:decidability}.
    \begin{applemma}\label{lem:slide-box-l}
        Let $ G \subseteq \agentSet $, with $ |G| \geq \ell $ and let $ \vec{v} \in G^* $ ($ |\vec{v}| = \lambda \geq \ell $) such that each agent in $G$ appears in $\vec{v}$ at least once. Let $ \rho $ and $ \tau $ be two permutations of elements of $ \vec{v} $. Then, for any $\varphi$, in the logic wC$_\ell$-S5$_n$ the following is a theorem:
        \begin{equation*}
            \B{\rho_1} \dots \B{\rho_\lambda} \varphi \leftrightarrow \B{\tau_1} \dots \B{\tau_\lambda} \varphi
        \end{equation*}

        \begin{proof}
            First, we notice that in the logic wC$_\ell$-S5$_n$, for any formula $\varphi$, the following formula is a theorem (recall that $\langle i_1, \dots, i_\ell \rangle$ is a sequence of agents with no repetitions, and that $\pi$ is a permutation of this sequence):
            \begin{equation}\label{eq:comm-l}
                \B{i_1} \dots \B{i_\ell}\varphi \leftrightarrow \B{\pi_{i_1}} \dots \B{\pi_{i_\ell}}\varphi
            \end{equation}

            \noindent This immediately follows from axiom \axiom{wC$_\ell$}.

            Second, by construction, we have that for each $ \rho_i $ there exists $ \tau_{k_i} $ such that $ \tau_{k_i} = \rho_i $. Consider $ \tau_{k_1} = \rho_1 $. Then, by iterating Equation \ref{eq:comm-l}, we obtain:
            \begin{align*}
                                &~ \B{\tau_1} \dots \overbrace{\left(\B{\tau_{{k_1}-1}} \B{\tau_{k_1}} \dots \B{\tau_{{k_1}+\ell-1}}\right)}^\ell \dots \B{\tau_\lambda} \varphi \\
                \leftrightarrow &~ \B{\tau_1} \dots            \left(\B{\tau_{k_1}} \B{\tau_{{k_1}-1}} \dots \B{\tau_{{k_1}+\ell-1}}\right)       \dots \B{\tau_\lambda} \varphi \\
                \dots           &~                                                                                                \\
                \leftrightarrow &~ \overbrace{\left(\B{\tau_1} \B{\tau_{k_1}} \dots \B{\tau_j}\right)}^\ell \dots \B{\tau_{{k_1}-1}} \dots \B{\tau_{{k_1}+\ell-1}} \dots \B{\tau_\lambda} \varphi \\
                \leftrightarrow &~ \B{\tau_{k_1}} \B{\tau_1} \dots \B{\tau_j} \dots \B{\tau_{{k_1}-1}} \dots \B{\tau_{{k_1}+\ell-1}} \dots \B{\tau_\lambda} \varphi
            \end{align*}
            By repeating this manipulation for $ \rho_2, \dots \rho_m $, we obtain the conclusion.
        \end{proof}
    \end{applemma}

    The following is the corresponding version of Lemma \ref{lem:ck-n} of Section \ref{sec:decidability}.
    \begin{applemma}\label{lem:ck-n-l}
        Let $ G = \{i_1, \dots, i_m\} \subseteq \agentSet $, with $ m \geq \ell $. In the logic wC$_\ell$-S5$_n$, for any $\varphi$ and $ \vec{v} \in G^* $ we have that $ \B{i_1} \dots \B{i_m} \varphi \rightarrow \B{v_1} \cdots \B{v_{|\vec{v}|}} \varphi $ is a theorem.

        \begin{proof}
            The proof is by induction on $|\vec{v}|$.
            For the base case ($|\vec{v}|=0$) we have that the formulae $ \B{i_h} \B{i_{h+1}} \dots \B{i_m} \varphi \rightarrow \B{i_{h+1}} \dots \B{i_m} \varphi $ ($ 1 \leq h < m $) and $ \B{i_m} \varphi \rightarrow \varphi $ are instances of \axiom{T}. Together with propositional reasoning, we get that $ \B{i_1} \dots \B{i_m} \varphi \rightarrow \varphi $ is a theorem.

            Let now $|\vec{v}| = \lambda$ and suppose, by inductive hypothesis, that $\B{i_1} \dots \B{i_m} \varphi \rightarrow \B{v_1} \cdots \B{v_{\lambda}} \varphi$ is a theorem (for any formula $\varphi$). We now show that, for each $ j \in G $, the formula $\B{i_1} \dots \B{i_m} \varphi \rightarrow \B{v_1} \cdots \B{v_{\lambda}} \B{j} \varphi$ is also a theorem. By inductive hypothesis, substituting $\varphi$ with $\B{j} \varphi$, the following is a theorem:
            \begin{equation*}
                \B{i_1} \dots \B{i_m} \B{j} \varphi \rightarrow \B{v_1} \cdots \B{v_{\lambda}} \B{j} \varphi.
            \end{equation*}

            \noindent Let $\rho$ be any permutation of $G = \{i_1, \dots, i_m\}$, such that $\rho_m = j$. By Lemma B.8, we can now rewrite the formula $\Box_{i_1} \dots \Box_{i_m}\Box_{j}\varphi$ as $\Box_{\rho_1} \dots \Box_{\rho_m}\Box_{j}\varphi$, which is $\Box_{\rho_1} \dots \Box_{\rho_m}\Box_{\rho_m}\varphi$. Then, we use Equation 2 as in the proof of Lemma 2 to rewrite the above formula as $\Box_{\rho_1} \dots \Box_{\rho_m}\varphi$. By using Lemma B.8, we can rewrite this formula as $\Box_{i_1} \dots \Box_{i_m}\varphi$.
            %
            Finally, we obtain that the following is a theorem:
            \begin{equation*}
                \B{i_1} \dots \B{i_m} \varphi \rightarrow \B{v_1} \cdots \B{v_{\lambda}} \B{j} \varphi.
            \end{equation*}
            This is the required result.
        \end{proof}
    \end{applemma}

    The following is the corresponding version of Theorem \ref{th:ck} of Section \ref{sec:decidability}.
    \begin{apptheorem}\label{th:ck-l}
        Let $ G = \{i_1, \dots, i_m\} \subseteq \agentSet $, with $ m \geq \ell $. In the logic wC$_\ell$-S5$_n$, for any $\varphi$, the formula $ \B{i_1} \dots \B{i_m} \varphi \leftrightarrow \CK{G} \varphi $ is a theorem.
        \begin{proof} 
            ($\Leftarrow$) This follows by definition of common knowledge; ($\Rightarrow$) this immediately follows by Lemma \ref{lem:ck-n-l}.
        \end{proof}
    \end{apptheorem}

    The following is the corresponding version of Corollary \ref{cor:diameter} of Section \ref{sec:decidability}.
    \begin{appcorollary}\label{cor:diameter-l}
        Let $ G {=} \{i_1, \dots, i_m\} \subseteq \agentSet $, with $ m \geq \ell $. 
        In an wC$_\ell$-S5$_n$-model, for any $\vec{v} \in G^*$, we have that if $w R_{v_1} \circ \ldots \circ R_{v_{|\vec{v}|}} w'$, then $w R_{i_1} \circ \dots \circ R_{i_m} w'$.
    \end{appcorollary}

    The statement above directly follows from the contrapositive of the implication in Lemma \ref{lem:ck-n-l}, under the assumption of minimality of states (w.r.t. bisimulation).
    Intuitively, this states that in a wC$_\ell$-S5$_n$-model, given any subset of $m \geq \ell$ agents, if a world is reachable in an arbitrary number of steps, then it is also reachable in exactly $m$ steps. Thus, in general, any pair of worlds of a wC$_\ell$-S5$_n$-model that are reachable one another are connected by a path of length \emph{at most $n$}.

    \settheoremcountertoref{th:dec-l}
    \begin{thm}\label{th:dec-l-proof}
        For any $n{>}1$ and $1 {<} \ell {\leq} n$, \planex{$\mathcal{T}_{\textnormal{wC$_\ell$-S5}}$}{$n$} is \emph{decidable}.
        \begin{proof}
            As a result of Corollary \ref{cor:diameter-l}, Lemmata \ref{lem:bounded-bisim} and \ref{lem:char-formulae} hold also in the logic wC$_\ell$-S5$_n$ (for any $\ell > 1$). Thus, as in the proof of Theorem \ref{th:dec}, let $T \in \mathcal{T}_{\textnormal{wC$_\ell$-S5}_n}$ be an epistemic planning task (for any such $\ell$). By Lemma \ref{lem:char-formulae}, it follows that we can perform a breadth-first search on the search space that would only visit a finite number of epistemic states (up to bisimulation contraction) to find a solution for $T$. Thus, we obtain the claim.
        \end{proof}
    \end{thm}

    \section{\nameref*{sec:systems}}
    First, we recall that axiom \axiom{C} is a Sahlqvist formula that corresponds to the following frame property on event models:
    \begin{equation}\label{eq:A-frame}
        \forall u, v, w (u Q_j v \wedge v Q_i w \rightarrow \exists x (u Q_i x \wedge x Q_j w))
    \end{equation}

    \begin{figure}[t]
        \centering{
        \begin{tikzpicture}
            \input{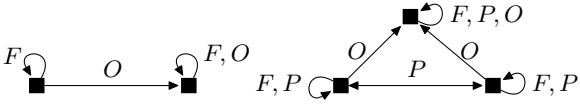}
        \end{tikzpicture}
        }
        \caption{Frames of S5$_n$-$m\mathcal{A}^*$ event models for ontic actions (left) and sensing/announcement actions (right).}
        \label{fig:ma_star_s5}
    \end{figure}

    \begin{figure}[t]
        \centering{
            \begin{tikzpicture}
                \input{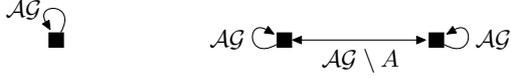}
            \end{tikzpicture}
        }
        \caption{Frames of event models in \cite{conf/aips/Kominis2015} for \emph{do} and \emph{update} actions (left) and \emph{sense} (right).}
        \label{fig:kom15}
    \end{figure}

    \setlemmacountertoref{lem:systems}
    \begin{lemma}\label{lem:systems-proof}
        $\mathcal{T}_{\textnormal{S5$_n$-}m\mathcal{A}^*} \subseteq \mathcal{T}_{\textnormal{C-S5}}$ and $\mathcal{T}_{\textnormal{KG}} \subseteq \mathcal{T}_{\textnormal{C-S5}}$.

        \begin{proof}
            We focus on S5$_n$-$m\mathcal{A}^*$ (the proof for the framework by Kominis and Geffner is analogous). It is easy to see that the frames of both public ontic actions (Figure \ref{fig:ma_star_s5}, left) and semi-private sensing/announcement actions (Figure \ref{fig:ma_star_s5}, right) are reflexive, symmetric and transitive.
            
            We now show that they both satisfy frame property (\ref{eq:A-frame}). As public ontic actions contain only one event, $e_1$, this kind of event model trivially satisfies frame property (\ref{eq:A-frame}). Thus, public ontic actions are C-S5$_n$-actions.

            We now move to semi-private sensing/announcement actions. We recall that, by construction in \cite{journals/corr/Baral2015}, we have that $F \cup P = \agentSet$ and $F \cap P = \varnothing$. Let $i,j \in \agentSet$. We now have four cases:
            \begin{enumerate}
                \item $i,j \in F$: we can only assign $u,v,w$ to either $u=v=w=f_1$ or $u=v=w=f_2$. In the former case, frame property (\ref{eq:A-frame}) is satisfied by choosing $x=f_1$ and, in the latter, by choosing $x=f_2$.
                \item $i,j \in P$: for any way of assigning $u,v,w$, frame property (\ref{eq:A-frame}) is satisfied by choosing $x=u$.
                \item $i \in F$ and $j \in P$: for any way of assigning $u,v,w$, frame property (\ref{eq:A-frame}) is satisfied by choosing $x=w$.
                \item $i \in P$ and $j \in F$: for any way of assigning $u,v,w$, frame property (\ref{eq:A-frame}) is satisfied by choosing $x=u$.
            \end{enumerate}

            This shows that the event models of semi-private sensing/announcement actions satisfy frame property (\ref{eq:A-frame}). Thus, semi-private sensing/announcement actions are C-S5$_n$-actions. This completes the proof.
        \end{proof}
    \end{lemma}


\begin{thebibliography}{10}
        \bibitem{conf/ijcai/Aucher2013}
        Guillaume Aucher and Thomas Bolander, `Undecidability in epistemic planning',
          in {\em {IJCAI} 2013, Proceedings of the 23rd International Joint Conference
          on Artificial Intelligence, Beijing, China, August 3-9, 2013}, ed., Francesca
          Rossi, pp. 27--33. {IJCAI/AAAI}, (2013).
        
        \bibitem{journals/corr/Aucher2014}
        Guillaume Aucher, Bastien Maubert, and Sophie Pinchinat, `Automata techniques
          for epistemic protocol synthesis', in {\em Proceedings 2nd International
          Workshop on Strategic Reasoning, {SR} 2014, Grenoble, France, April 5-6,
          2014}, eds., Fabio Mogavero, Aniello Murano, and Moshe~Y. Vardi, volume 146
          of {\em {EPTCS}}, pp. 97--103, (2014).
        
        \bibitem{journals/corr/Baral2015}
        Chitta Baral, Gregory Gelfond, Enrico Pontelli, and Tran~Cao Son, `An action
          language for multi-agent domains: Foundations', {\em CoRR}, {\bf
          abs/1511.01960}, (2015).
        
        \bibitem{books/cup/Blackburn2001}
        Patrick Blackburn, Maarten~de Rijke, and Yde Venema, {\em Modal Logic},
          Cambridge Tracts in Theoretical Computer Science, Cambridge University Press,
          2001.
        
        \bibitem{journals/jancl/Bolander2011}
        Thomas Bolander and Mikkel~Birkegaard Andersen, `Epistemic planning for single
          and multi-agent systems', {\em J. Appl. Non Class. Logics}, {\bf 21}(1),
          9--34, (2011).
        
        \bibitem{journals/ai/Bolander2020}
        Thomas Bolander, Tristan Charrier, Sophie Pinchinat, and Fran{\c{c}}ois
          Schwarzentruber, `Del-based epistemic planning: Decidability and complexity',
          {\em Artif. Intell.}, {\bf 287},  103304, (2020).
        
        \bibitem{conf/kr/Bolander2021}
        Thomas Bolander, Lasse Dissing, and Nicolai Herrmann, `{DEL-based Epistemic
          Planning for Human-Robot Collaboration: Theory and Implementation}', in {\em
          {Proceedings of the 18th International Conference on Principles of Knowledge
          Representation and Reasoning}}, pp. 120--129, (11 2021).
        
        \bibitem{conf/ijcai/Bolander2015}
        Thomas Bolander, Martin~Holm Jensen, and Fran{\c{c}}ois Schwarzentruber,
          `Complexity results in epistemic planning', in {\em Proceedings of the
          Twenty-Fourth International Joint Conference on Artificial Intelligence,
          {IJCAI} 2015, Buenos Aires, Argentina, July 25-31, 2015}, eds., Qiang Yang
          and Michael~J. Wooldridge, pp. 2791--2797. {AAAI} Press, (2015).
        
        \bibitem{conf/ijcai/Charrier2016}
        Tristan Charrier, Bastien Maubert, and Fran{\c{c}}ois Schwarzentruber, `On the
          impact of modal depth in epistemic planning', in {\em Proceedings of the
          Twenty-Fifth International Joint Conference on Artificial Intelligence,
          {IJCAI} 2016, New York, NY, USA, 9-15 July 2016}, ed., Subbarao Kambhampati,
          pp. 1030--1036. {IJCAI/AAAI} Press, (2016).
        
        \bibitem{books/mit/Fagin2004}
        Ronald Fagin, Joseph~Y. Halpern, Yoram Moses, and Moshe~Y. Vardi, {\em
          Reasoning About Knowledge}, {MIT} Press, 2004.
        
        \bibitem{book/nhpc/Gabbay2003}
        D.M. Gabbay, {\em Many-dimensional Modal Logics: Theory and Applications},
          Studies in logic and the foundations of mathematics, North Holland Publishing
          Company, 2003.
        
        \bibitem{books/el/Goranko2007}
        Valentin Goranko and Martin Otto, `Model theory of modal logic', in {\em
          Handbook of Modal Logic}, eds., Patrick Blackburn, J.~F. A.~K. van Benthem,
          and Frank Wolter, volume~3 of {\em Studies in logic and practical reasoning},
           249--329, North-Holland, (2007).
        
        \bibitem{book/springer/Gray1978}
        Jim Gray, `Notes on data base operating systems', in {\em Operating Systems, An
          Advanced Course}, pp. 393--481, Berlin, Heidelberg, (1978). Springer-Verlag.
        
        \bibitem{journals/jacm/Halpern1990}
        Joseph~Y. Halpern and Yoram Moses, `Knowledge and common knowledge in a
          distributed environment', {\em J. {ACM}}, {\bf 37}(3),  549--587, (1990).
        
        \bibitem{workshop/nrac/Herzig2005}
        Andreas Herzig, Paul Sabatier, J{\'e}r{\^o}me Lang, and Pierre Marquis, `Action
          progression and revision in multiagent belief structures', in {\em Sixth
          Workshop on Nonmonotonic Reasoning, Action, and Change ({NRAC})}, (2005).
        
        \bibitem{phd/dtu/Jensen2014}
        Martin~Holm Jensen, {\em {Epistemic and Doxastic Planning}}, Ph.D.\
          dissertation, Technical University of Denmark, 2014.
        
        \bibitem{conf/aips/Kominis2015}
        Filippos Kominis and Hector Geffner, `Beliefs in multiagent planning: From one
          agent to many', in {\em Proceedings of the Twenty-Fifth International
          Conference on Automated Planning and Scheduling, {ICAPS} 2015, Jerusalem,
          Israel, June 7-11, 2015}, eds., Ronen~I. Brafman, Carmel Domshlak, Patrik
          Haslum, and Shlomo Zilberstein, pp. 147--155. {AAAI} Press, (2015).
        
        \bibitem{conf/lori/Lowe2011}
        Benedikt L{\"{o}}we, Eric Pacuit, and Andreas Witzel, `{DEL} planning and some
          tractable cases', in {\em Logic, Rationality, and Interaction - Third
          International Workshop, {LORI} 2011, Guangzhou, China, October 10-13, 2011.
          Proceedings}, eds., Hans van Ditmarsch, J{\'{e}}r{\^{o}}me Lang, and Shier
          Ju, volume 6953 of {\em Lecture Notes in Computer Science}, pp. 179--192.
          Springer, (2011).
        
        \bibitem{book/ph/Minsky1967}
        Marvin~L Minsky, {\em Computation: finite and infinite machines},
          Prentice-Hall, Inc., 1967.
        
        \bibitem{journals/synthese/Paternotte11}
        C{\'{e}}dric Paternotte, `Being realistic about common knowledge: a lewisian
          approach', {\em Synth.}, {\bf 183}(2),  249--276, (2011).
        
        \bibitem{conf/aaai/Son2015}
        Tran~Cao Son, Enrico Pontelli, Chitta Baral, and Gregory Gelfond, `Exploring
          the {KD45} property of a kripke model after the execution of an action
          sequence', in {\em Proceedings of the Twenty-Ninth {AAAI} Conference on
          Artificial Intelligence, January 25-30, 2015, Austin, Texas, {USA}}, eds.,
          Blai Bonet and Sven Koenig, pp. 1604--1610. {AAAI} Press, (2015).
        
        \bibitem{journals/csur/Torreno2017}
        Alejandro Torre{\~{n}}o, Eva Onaindia, Anton{\'{\i}}n Komenda, and Michal
          Stolba, `Cooperative multi-agent planning: {A} survey', {\em {ACM} Comput.
          Surv.}, {\bf 50}(6),  84:1--84:32, (2018).
        
        \bibitem{book/springer/vanDitmarsch2007}
        Hans~P. van Ditmarsch, Wiebe van~der Hoek, and Barteld~P. Kooi, {\em Dynamic
          Epistemic Logic}, volume 337, Springer Netherlands, 2007.
        
        \bibitem{conf/ijcai/Yu2013}
        Quan Yu, Ximing Wen, and Yongmei Liu, `Multi-agent epistemic explanatory
          diagnosis via reasoning about actions', in {\em {IJCAI} 2013, Proceedings of
          the 23rd International Joint Conference on Artificial Intelligence, Beijing,
          China, August 3-9, 2013}, ed., Francesca Rossi, pp. 1183--1190. {IJCAI/AAAI},
          (2013).
        
    \end{thebibliography}

\end{document}